\documentclass[10pt,journal,compsoc]{IEEEtran}
\ifCLASSOPTIONcompsoc
  \usepackage[nocompress]{cite}
\else
  \usepackage{cite}
\fi
\usepackage{color, colortbl}
\definecolor{mygray}{gray}{0.9}

\usepackage{xcolor}
\usepackage{bm}
\usepackage{mathrsfs}  

\usepackage{epsfig}
\usepackage{graphicx}
\usepackage{amsmath}
\usepackage{amssymb}
\usepackage{helvet}  
\usepackage{courier}  
\usepackage{url}  
\usepackage{multirow}
\usepackage{subcaption}
\usepackage{array}
\usepackage[export]{adjustbox}
\usepackage{hyperref}
\usepackage{booktabs}
\hypersetup{
    colorlinks=true,
    linkcolor=blue,
    filecolor=magenta,      
    urlcolor=cyan,
}
\ifCLASSINFOpdf
\else
\fi
\begin{document}
%
\title{Conditional Modeling Based Automatic Video Summarization}


%
%
%
%

\author{Jia-Hong Huang,~\IEEEmembership{Member,~IEEE,}
        Chao-Han Huck Yang,~\IEEEmembership{Member,~IEEE,}
        Pin-Yu Chen,~\IEEEmembership{Member,~IEEE,}
        Min-Hung Chen,~\IEEEmembership{Member,~IEEE,}
        and~Marcel Worring,~\IEEEmembership{Senior Member,~IEEE}
\thanks{Manuscript received November 20, 2023; revised March 20, 2024.}}

%
%

\markboth{Journal of \LaTeX\ Class Files,~Vol.~14, No.~8, November~2023}%
{Shell \MakeLowercase{\textit{et al.}}: Bare Demo of IEEEtran.cls for Computer Society Journals}
%



\IEEEtitleabstractindextext{%
\begin{abstract}
The aim of video summarization is to shorten videos automatically while retaining the key information necessary to convey the overall story. Video summarization methods mainly rely on visual factors, such as visual consecutiveness and diversity, which may not be sufficient to fully understand the content of the video. There are other non-visual factors, such as interestingness, representativeness, and storyline consistency that should also be considered for generating high-quality video summaries. Current methods do not adequately take into account these non-visual factors, resulting in suboptimal performance. In this work, a new approach to video summarization is proposed based on insights gained from how humans create ground truth video summaries. The method utilizes a conditional modeling perspective and introduces multiple meaningful random variables and joint distributions to characterize the key components of video summarization. Helper distributions are employed to improve the training of the model. A conditional attention module is designed to mitigate potential performance degradation in the presence of multi-modal input. The proposed video summarization method incorporates the above innovative design choices that aim to narrow the gap between human-generated and machine-generated video summaries. Extensive experiments show that the proposed approach outperforms existing methods and achieves state-of-the-art performance on commonly used video summarization datasets.
\end{abstract}

}

\maketitle

\IEEEdisplaynontitleabstractindextext

%
\IEEEpeerreviewmaketitle


\IEEEraisesectionheading{\section{Introduction}\label{sec:introduction}}
Video summarization plays a crucial role in enhancing video browsing, searchability, and comprehension, empowering users to efficiently navigate through video collections and extract relevant information. A good summary is a video clip that is short and conveys the key message and narrative of the original video. To achieve this, there has been a recent surge of interest in developing automatic video summarization algorithms, \cite{plummer2017enhancing,chu2015video,panda2017collaborative,potapov2014category,rochan2019video,li2018local,zhou2018deep,sharghi2018improving,zhang2018retrospective,wu2022intentvizor}.

\begin{figure*}[t!]
\begin{center}
\includegraphics[width=1.0\linewidth,height=6.5cm]{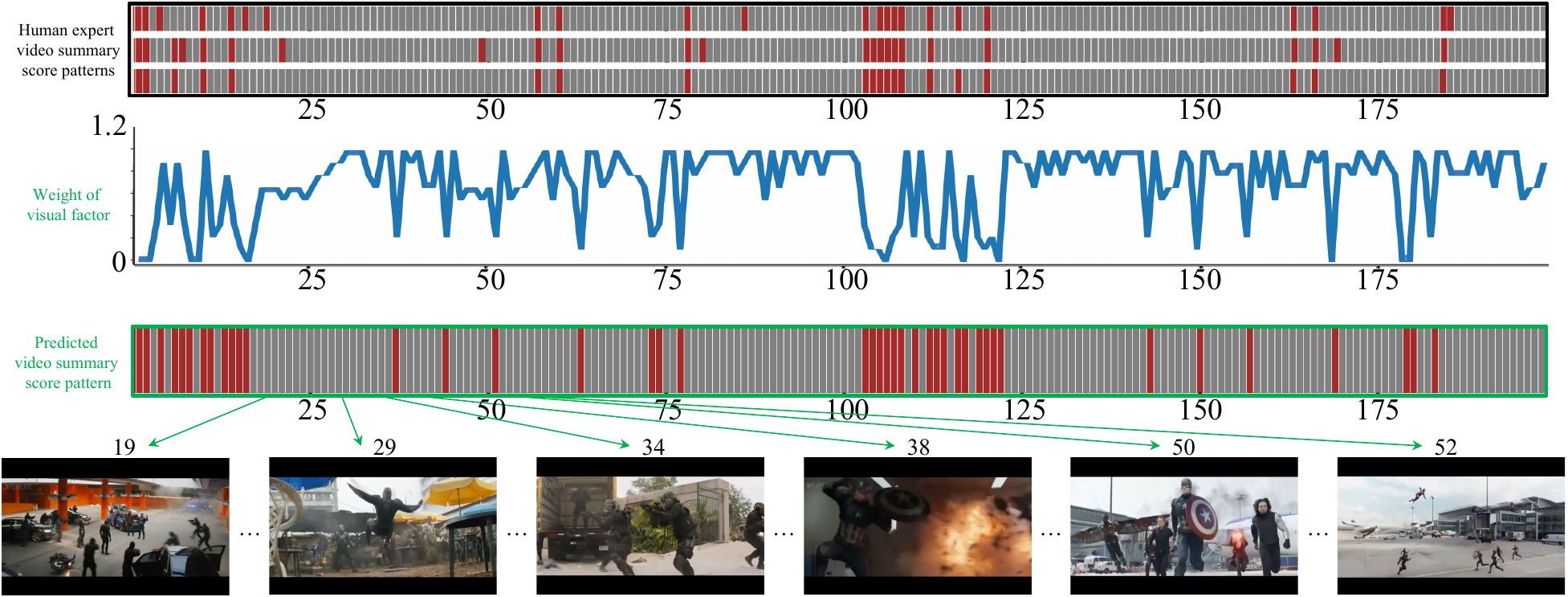}
\end{center}
    \captionof{figure}{Visualization of human-annotated and machine-predicted frame-level scores used for creating video summaries. Upon comparing the score patterns of human-annotated video summaries to those generated by current state-of-the-art video summarization methods, e.g., \cite{vasudevan2017query,huang2020query,huang2021gpt2mvs}, it becomes apparent that these methods are able to learn the importance of visual consecutiveness and diversity, which are crucial factors in creating effective video summaries, as also valued by humans. The focus of these methods is primarily on capturing the visual factors to achieve these goals. When comparing the three score patterns of human-annotated video summaries of the same video, we can observe that the three patterns are different from each other. This implies that when creating a video summary, humans not only consider visual factors but also take into account non-visual factors. Discarded frames are marked in red bars while grey bars represent the selected frames used for constructing the video summary. The video contains $199$ frames, and the numbers indicate the frame indices. The term "Weight of visual factor" is defined as the proportion or percentage of common visual objects present in a frame. 
    }
\label{fig:figure21}
\end{figure*}

When comparing the automatic algorithms with human experts, as suggested by \cite{gygli2015video,song2015tvsum,vasudevan2017query,gong2014diverse,gygli2014creating,huang2020query}, it becomes evident that human experts take into account a variety of factors when creating video summaries. These factors encompass both concrete/visual factors such as visual consecutiveness and visual diversity and abstract/non-visual factors such as interestingness, representativeness, and storyline consistency. These factors have a significant impact on the final outcome of the video summary and must be taken into account to ensure the quality of the summary. However, current video summarization methods rely primarily on visual factors, as illustrated in Figure \ref{fig:figure21}, and tend to not, or in a very limited way, consider non-visual factors. Not taking into account non-visual factors leads to suboptimal performance of automatic video summarization \cite{huang2020query,jiang2022joint}. 


\begin{figure}[t!]
\begin{center}
\includegraphics[width=1.0\linewidth]{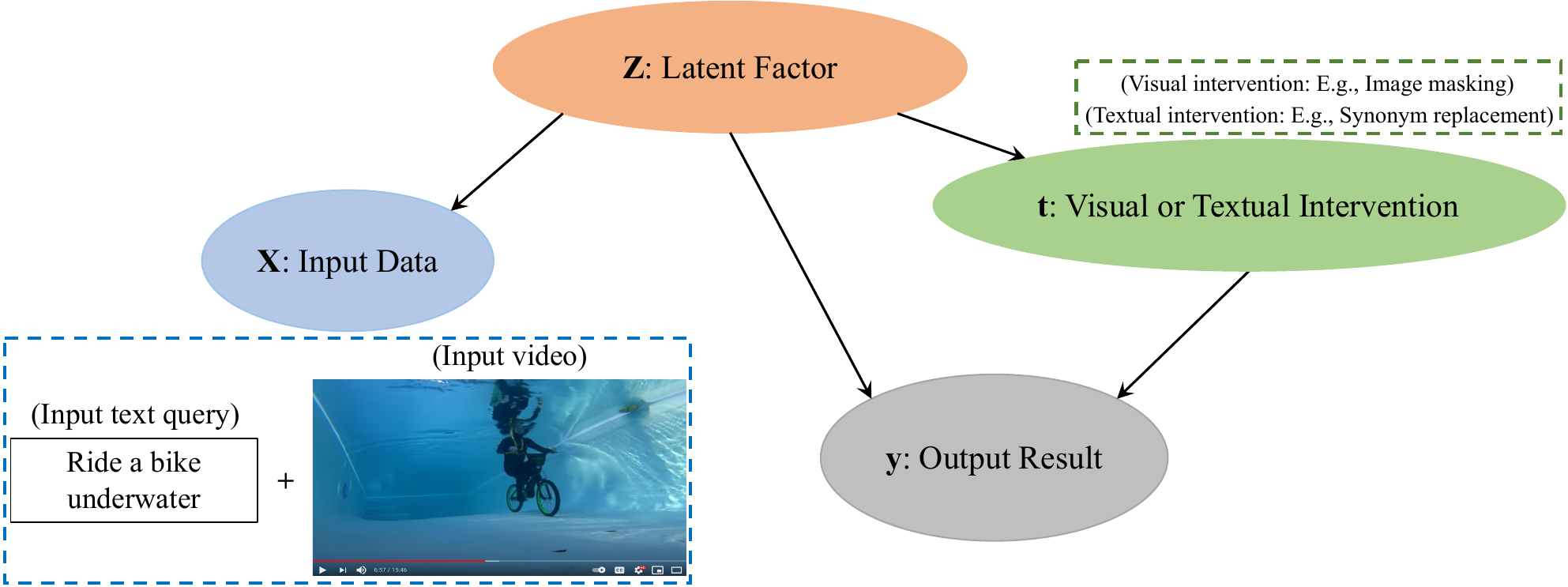}
\end{center}
\vspace{-0.50cm}
  \caption{Example of a conditional graph in video summarization. $\textbf{t}$ is an intervention, e.g., visual or textual perturbation. $\textbf{y}$ is an outcome, e.g., an importance score of a video frame or a relevance score between the input text query and video. $\textbf{Z}$ is an unobserved factor, e.g., representativeness, interestingness, or storyline smoothness. $\textbf{X}$ is containing noisy views on the hidden factor $\textbf{Z}$, say the input text query and video. The conditional graph of video summarization leads to more explainable modeling.
  }
\label{fig:figure39}
\end{figure}

In this work we start from the premise that the relation between the human driven non-visual factors and the final summary is complex and non-deterministic. We therefore study the problem from a conditional modeling perspective. To create a model, we incorporate insights derived from \cite{song2015tvsum,gygli2014creating}, which investigate how humans evaluate the quality of video summaries, as illustrated in Figure \ref{fig:figure39}. The foundation for our approach lies in data intervention
\cite{louizos2017causal}. 
Researchers have drawn inspiration from intervention modeling techniques \cite{greenland1983correcting,selen1986adjusting} and incorporated proxy variables to explore the conditions for conditional learning on large-scale datasets \cite{cai2012identifying,greenland2011bias,louizos2017causal,shalit2017estimating}. Such methods enhance the model's ability to learn relationships by introducing additional information to the learning process.
Our proposed video summarization approach incorporates four significant and meaningful random variables, which for video summarization characterize the behavior of data intervention \cite{louizos2017causal}, the model's prediction, observed factors, and unobserved factors, respectively.
To model the insights on human summary assessment, the proposed method involves building a prior joint distribution and its posterior approximation, based on the aforementioned four random variables. 

\begin{figure*}[ht]
\begin{center}
\includegraphics[width=1.0\linewidth]{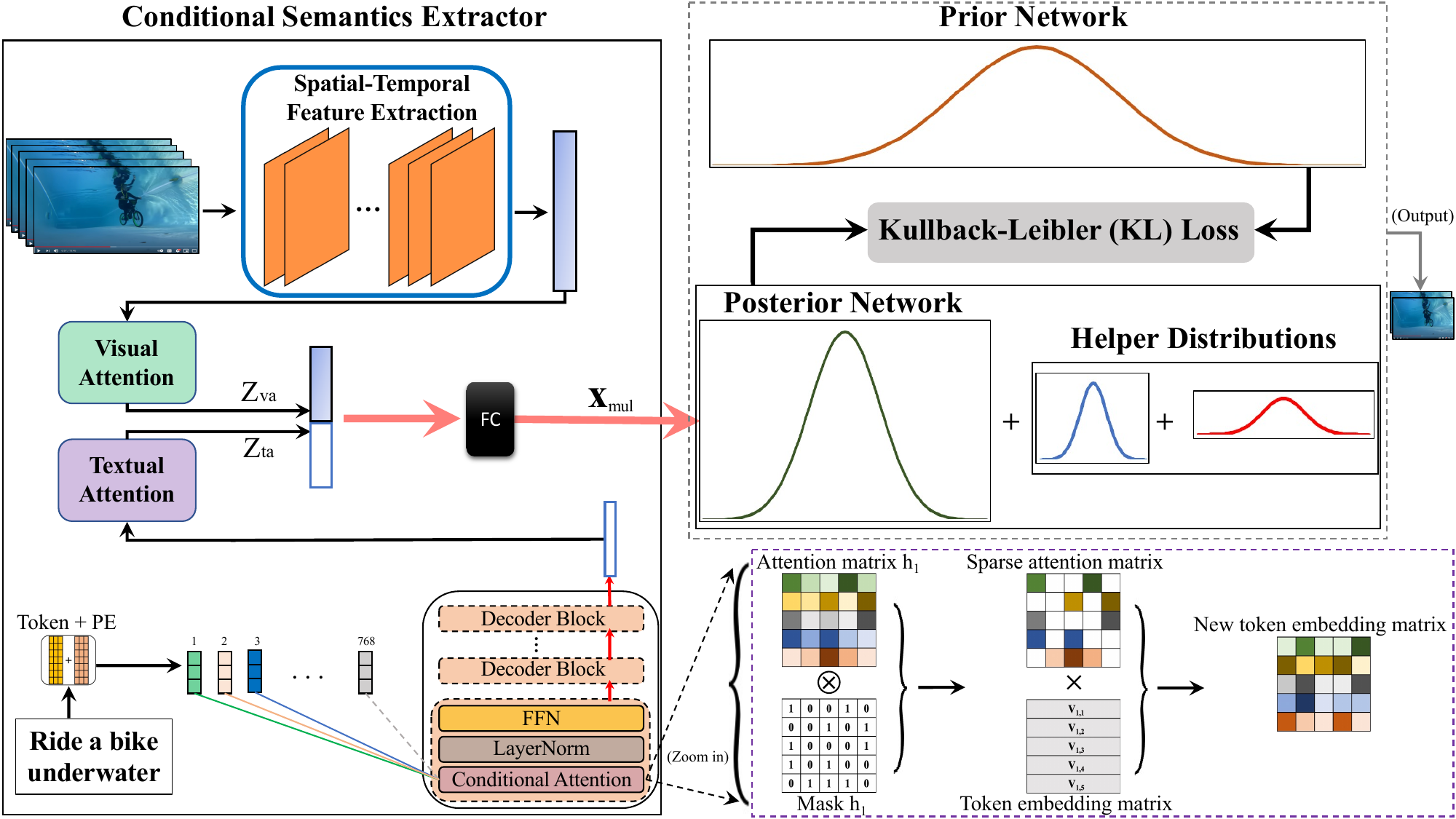}
\end{center}
\vspace{-0.50cm}
   \caption{Flowchart of the proposed method for video summarization. The proposed method is mainly composed of a prior network, a posterior network, helper distributions, and a conditional attention module. $\otimes$ denotes element-wise multiplication and $\times$ indicates matrix multiplication. ``Token + PE'' denotes the operations of token embedding and positional encoding.
   }
\label{fig:figure0}
\end{figure*}

The method is trained by minimizing the distance between the prior distribution and the posterior approximation. However, predicting the behaviors of the data intervention and the model outcome can be challenging in practice due to various factors like video noise or motion/lens blur. To address this, helper distributions are introduced, and a new loss term is created to guide model learning. When the visual data is accompanied by textual information, the extra input can sometimes harm the model's performance due to ineffective interactions between different modalities. To overcome this challenge, a conditional attention module is introduced to effectively distill the mutual information between the multi-modal inputs. Figure \ref{fig:figure0} illustrates the flowchart of the proposed approach. The proposed video summarization method incorporates the above innovative design choices, i.e., conditional modeling, helper distributions, and conditional attention module, aimed at minimizing the gap between human-generated and machine-generated video summaries. Extensive experiments conducted on commonly used video summarization datasets demonstrate that the proposed approach outperforms existing methods and achieves state-of-the-art performance.

\vspace{+3pt}
\noindent\textbf{Contributions.}
\vspace{-0.1cm}
\begin{itemize}
    \item A novel approach to model the video summarization problem, inspired by the way humans approach the task.
    

    \item To effectively extract the interactive information between multi-modal inputs, a conditional attention module is introduced in scenarios where there are multiple modes of input.
    
    \item The effectiveness of the proposed method is verified through extensive experiments on widely used video summarization benchmark datasets. The experimental results indicate that the proposed approach is successful and surpasses other methods, achieving state-of-the-art performance in terms of $F_{1}$-score. 
\end{itemize}

The remainder of this paper is structured as follows: In Section 2, we provide a review of related works. Section 3 presents the details of our proposed method, including its design choices and implementation. In Section 4, we evaluate the performance of our approach and compare it to existing state-of-the-art methods. Finally, in Section 5 and in Section 6, we discuss our findings and highlight future research directions.

\noindent
\vspace{+0.1cm}\textbf{Relations to our previous work}

This paper represents an advancement over our previous conference paper that was presented at the International Conference on Multimedia and Expo (ICME) in 2022 \cite{huang2022causal}. 
Firstly, we introduce a conditional attention module (see Figure \ref{fig:figure0}) that effectively extracts the interaction between multi-modal inputs. Secondly, we perform comprehensive and extended experiments using TVSum \cite{song2015tvsum}, SumMe \cite{gygli2014creating}, and QueryVS \cite{huang2020query} datasets. Thirdly, we demonstrate how to incorporate visual and textual perturbations as data interventions in the conditional modeling approach for video summarization. Furthermore, the current paper is a fully restructured and rewritten version of the previous work.

\section{Related Work}
In the following section, we provide a brief review of relevant works that are related to the proposed approach. Video summarization is a machine learning challenge that can be approached with different supervision schemes, such as fully-supervised, weakly-supervised, or unsupervised methods. The subsequent sections will provide a concise summary of relevant methods in each category.


\subsection{Fully-supervised Methods with Visual Input Only}

Video summarization commonly involves fully-supervised learning \cite{gong2014diverse,gygli2014creating,zhang2016video,zhao2017hierarchical,zhao2018hsa,zhang2019dtr,ji2019video,ji2020deep}, where human-defined labels are used to supervise the training phase. These methods can be categorized into three classes:
The first class involves Recurrent Neural Network (RNN)/Long Short-Term Memory (LSTM) approaches \cite{hochreiter1997long,zhao2017hierarchical,zhao2018hsa,zhang2019dtr,ji2019video}. In \cite{zhao2017hierarchical,zhao2018hsa}, RNN architectures are employed in a hierarchical manner to model temporal structures in video data and select shots/segments for the summary. In \cite{zhang2019dtr}, a dilated temporal relational generative adversarial network is introduced, which captures temporal dependencies with LSTM and dilated temporal relational units to address frame-based video summarization. \cite{ji2019video} treats video summarization as a sequence-to-sequence learning problem, proposing an LSTM-based encoder-decoder model with an intermediate attention layer. This approach was later extended by integrating a semantics preserving embedding network \cite{ji2020deep}.
The second class involves the use of the Determinantal Point Process (DPP) \cite{kulesza2012determinantal,zhang2016video}. In \cite{zhang2016video}, video summarization is treated as a structured prediction problem, and a deep-learning-based approach is proposed to estimate the importance of video frames. LSTM is used to model temporal dependencies, and the DPP is employed to increase content diversity.
The third class comprises methods that do not involve RNN/LSTM and DPP \cite{gong2014diverse,gygli2014creating}. In \cite{gong2014diverse}, a probabilistic model called the sequential DPP is introduced to capture the sequential structures of video data and select diverse subsets. The sequential DPP acknowledges the intrinsic sequential structures present in video data, thereby addressing the limitation of the standard DPP, which treats video frames as randomly permutable entities.\cite{gygli2014creating} proposes an automatic summarization method for user videos with interesting events, utilizing features to predict visual interestingness and selecting an optimal set of superframes for the summary. \\
The fully-supervised approaches mentioned above rely on complete human expert annotations for training the model, which indeed leads to excellent performance. However, obtaining such annotations can be costly. Therefore, there is a need for a more cost-effective solution for video summarization. In this work, we propose such a more cost-effective alternative method based on conditional modeling. This approach offers improved interpretability, better generalization, increased flexibility, and enhanced decision-making compared to existing methods. 

\subsection{Fully-supervised Methods with Multi-modal Input}
Several approaches have been proposed to enhance video summarization by exploiting additional modalities beyond the visual input. These modalities include viewers' comments, video captions, or other contextual data \cite{li2017extracting,vasudevan2017query,sanabria2019deep,song2016category,zhou2018video,lei2018action,otani2016video,yuan2017video,wei2018video,huang2020query,huang2021gpt2mvs,huang2022causal,huang2023causalainer,huang2023query,hu2019silco,huck2018auto,liu2018synthesizing,yang2018novel,di2021dawn,wu2023expert}. For instance, a multi-modal video summarization approach is introduced in \cite{li2017extracting} to extract key-frames from first-person videos. In \cite{sanabria2019deep}, a deep-learning-based method is proposed to summarize soccer game videos using multiple modalities. \cite{song2016category} introduces a model for category-driven video summarization that preserves core parts found in summaries of the same category \cite{zhou2018video}. A reinforcement learning-based approach is proposed in \cite{lei2018action} that uses action classifiers trained with video-level labels for action-driven video fragmentation and labeling, followed by category-driven video summarization. In \cite{otani2016video,yuan2017video}, video summaries are generated by maximizing their relevance with the available metadata of the video, with visual and textual information projected onto a common latent feature space. Finally, in \cite{wei2018video}, a visual-to-text mapping and a semantic-based selection of video fragments are applied based on the match between the automatically generated and the original video descriptions, using semantic attended networks. \\
Multi-modal approaches, such as the above, for video summarization utilize additional modalities to improve the performance of the model. Effective fusion of these modalities is crucial for the success of the approach. The utilization of inadequate fusion methods, as mentioned in \cite{huang2020query,vasudevan2017query,wei2018video}, can limit the model's capability to leverage the complementary information provided by different modalities to its fullest potential.

\subsection{Weakly-supervised Methods}
Several video summarization approaches, such as those in \cite{panda2017weakly,ho2018summarizing,cai2018weakly,chen2019weakly,jiang2019comprehensive,yan2020self}, adopt weakly-supervised learning to overcome the need for extensive data with human expert annotations. These methods leverage less-expensive weak labels, such as video-level annotations from human experts, to train models. Although weak labels are less accurate than full human expert annotations, they can still effectively train video summarization models and achieve acceptable performance. For instance, in \cite{panda2017weakly}, a weakly-supervised approach is introduced that adopts an intermediate method between fully-supervised learning and unsupervised learning. It uses video-level metadata to categorize videos, extract 3D-CNN features, and learn a video summarization model for categorizing new videos. Similarly, in \cite{cai2018weakly}, a weakly-supervised approach uses a model structure that combines the architectures of the encoder-decoder with a soft attention mechanism and the Variational AutoEncoder (VAE) to learn latent semantics from web videos. The proposed video summarization model is trained with a weakly-supervised semantic matching loss to generate video summaries. In \cite{ho2018summarizing}, a model is proposed that trains on fully-annotated highlight scores from third-person videos and a small set of annotated first-person videos. Note that, in this case, only a small portion of the set of first-person videos comes with ground-truth annotations. In \cite{chen2019weakly}, reinforcement learning is used to train a video summarization model with a limited set of human annotations and handcrafted rewards. The proposed method applies a hierarchical key-fragment selection process and generates a final video summary based on rewards for diversity and representativeness. \\
The above weakly-supervised video summarization methods
are more cost-effective than fully-supervised approaches; however, they may not always lead to optimal performance compared to the latter.

\subsection{Unsupervised Methods}
Unsupervised methods for video summarization rely on the idea that a good summary should 
convey the essence of the video and that the evidence for that should be present in the data. Thus, we would only need ways to capture this information \cite{zhao2014quasi,chu2015video,panda2017collaborative,mahasseni2017unsupervised,rochan2019video,herranz2012scalable,apostolidis2019stepwise,jung2019discriminative,yuan2019cycle,apostolidis2020unsupervised,sheinfeld2016video}. In \cite{zhao2014quasi}, a dictionary is learned from the video data using group sparse coding, and a summary is created by combining segments that cannot be reconstructed sparsely based on the dictionary. \cite{chu2015video} introduces a maximal biclique finding algorithm to identify frequently co-occurring visual patterns and generates a summary by selecting shots that co-occur most often. A video summarization model that simultaneously captures specific and general features is presented in \cite{panda2017collaborative}. \cite{mahasseni2017unsupervised} proposes a model that combines an LSTM-based key-frame selector, a trainable discriminator, and a VAE to create summaries through adversarial learning. \cite{apostolidis2019stepwise} proposes a stepwise label-based approach to train the adversarial part of the network proposed in \cite{mahasseni2017unsupervised}. In \cite{jung2019discriminative}, a VAE-GAN architecture is proposed, and in \cite{yuan2019cycle}, a cycle-consistent adversarial learning objective is used to maximize the mutual information between the summary and the video. A variation of \cite{apostolidis2019stepwise} replaces the VAE with a deterministic attention auto-encoder in \cite{apostolidis2020unsupervised}, leading to more effective key-fragment selection. \cite{rochan2019video} introduces a new formulation for summarizing video data, enforcing a diversity constraint on the summary mapping through adversarial training. \\
Existing unsupervised methods do not rely on human expert annotations or pseudo labels for supervision during the training phase. Consequently, their performance often falls behind that of fully-supervised methods. In contrast, our proposed method leverages both human expert annotations and pseudo-label supervision to gain an advantage.
It aims to reason about cause-effect relationships between input data and desired summaries, which is often overlooked in current machine learning-based video summarization methods. Our approach is versatile and can be applied to both fully-supervised and weakly-supervised schemes, based on different experimental setups. 

\section{Methodology}
\label{methodology:method}
Video summarization is the process of automatically creating a condensed representation of a video, capturing the most important and informative content from the original video. 
Given an input video $\textbf{x}=(x_{0},x_{1},..., x_{n})$ composed of $n$ frames we view video summarization as the problem of finding a classification function $f : y_{i}=f(x_{i})$ which determines whether $x_{i}$ is relevant for the summary $y_{i} = 1$ or irrelevant $y_{i}=0$. The classification problem is constraint by the user defined summary budget $N_{Y}$ such that $\sum_{i=0}^{n} y_{i} \leq N_{Y}$. Note that the summary budget is considered as a user-defined hyper-parameter \cite{huang2020query}.
In this study, our objective is to model video summarization as a conditional learning problem. Our proposed approach leverages conditional modeling to enhance the conditional inference ability of a machine learning-based video summarization model.
Let's delve into the components involved. We have the proxy variables $\textbf{X}$ and the hidden factor variable  $\textbf{Z}$. In the process of assimilating information from noisy interventions \cite{cai2012identifying,greenland2011bias,louizos2017causal,miao2018identifying,pearl2012measurement,kuroki2014measurement,wooldridge2009estimating,edwards2015all}, the model's first task is to heighten its inferential capabilities regarding the joint distribution $p(\textbf{X}, \textbf{Z})$ encompassing these proxy and hidden variables. Subsequently, armed with this enhanced understanding, the model adeptly employs this knowledge to fine-tune and adjust the hidden factor variable.
We begin by presenting the assumptions of conditional modeling. Then, we introduce four random variables: $\textbf{y}$, $\textbf{t}$, $\textbf{X}$, and $\textbf{Z}$, which respectively characterize the model’s prediction behavior, data intervention, and observed and unobserved factors. Specifically, from a modeling perspective of video summarization, the random variable $\textbf{X}$ represents an input video with or without a text query. The variable $\textbf{t}$ denotes a visual or textual intervention assignment, and $\textbf{y}$ represents the relevance score between the input text-based query and a video frame, or an importance score of a video frame. The video summary is created based on $\textbf{y}$. 
Afterward, we present the derivation of our training objective with helper distributions and the proposed conditional attention module. The proposed approach is mainly composed of two probabilistic networks: the prior and posterior networks. Please refer to Figure \ref{fig:figure0} for an overview. 

\subsection{Assumptions}

In general, conditional modeling for real-world observational studies is complicated \cite{louizos2017causal,abbasnejad2020counterfactual,yang2021causal,agarwal2020towards}. With the established efforts \cite{pearl2018theoretical,louizos2017causal,zhang2018advances} on conditional modeling under noisy interventions, two assumptions are imposed when modeling the problem of video summarization. 
First, the information of having visual/textual intervention $\textbf{t}$ or not is binary. Second, the observations $(\textbf{X}, \textbf{t}, \textbf{y})$ from a deep neural network (DNN) are sufficient to approximately recover the joint distribution $p(\textbf{Z}, \textbf{X}, \textbf{t}, \textbf{y})$ of the unobserved factor variable $\textbf{Z}$, the observed factor variable $\textbf{X}$, the intervention $\textbf{t}$, and the outcome $\textbf{y}$. The proposed method is built on top of multiple probability distributions as described in the following subsections.

\subsection{Conditional Modeling for Video Summarization}

To train our model we assume that we have a video collection at our disposal. Now let $\textbf{x}_i$ denote an input video and an optional text-based query indexed by $i$ in this collection, $\textbf{z}_i$ indicates the latent factor, $t_i \in \{0,1\}$ denotes the intervention assignment, and $y_i$ indicates the outcome. 

\noindent\textbf{Prior Probability Distributions.}
The prior network is conditioning on the latent variable $\textbf{z}_i$ and mainly consists of the following components: \\
\noindent(i) The latent factor distribution: 
\begin{align}
    p(\textbf{z}_i) = \prod_{z\in \textbf{z}_i} \mathcal{N}(z | \mu=0, \sigma^2=1),
\label{eq:eq1}
\end{align}
where $\mathcal{N}(\cdot | \mu, \sigma^2)$ denotes a Gaussian distribution with a random variable $z$, $z$ is an element of $\textbf{z}_i$, and the mean $\mu$ and variance $\sigma^2$ where we follow the settings in \cite{kingma2013auto}, i.e., $\mu=0$ $\sigma^2=1$. \\
\noindent(ii) The conditional data distribution:
\begin{align}
    p(\textbf{x}_i | \textbf{z}_i) = \prod_{x \in \ \textbf{x}_i} p(x | \textbf{z}_i),
\label{eq:eq2}
\end{align}
where $p(x |\textbf{z}_i)$ is an appropriate probability distribution with a random variable $x$, the distribution is conditioning on $\textbf{z}_i$, and $x$ is an element of $\textbf{x}_i$. \\
\noindent(iii) The conditional intervention distribution:
\begin{align}
    p(t_i | \textbf{z}_i) = \textup{Bernoulli} (\sigma(f_{\theta_1}(\textbf{z}_i))),
\label{eq:eq3}
\end{align}
where $\sigma(\cdot)$ is a logistic function, $\textup{Bernoulli}  (\cdot)$ indicates a Bernoulli distribution for a discrete outcome, and $f_{\theta_1}(\cdot)$ denotes a neural network parameterized by the parameter $\theta_1$. \\
\noindent(iv) The conditional outcome distribution:
\begin{align}
    p(y_i | \textbf{z}_i, t_i) = \sigma(t_{i} f_{\theta_2}(\textbf{z}_i) + (1 - t_{i}) f_{\theta_3}(\textbf{z}_i)),
\label{eq:eq4}
\end{align}
where $f_{\theta_2}(\cdot)$ and $f_{\theta_3}(\cdot)$ are neural networks parameterized by parameter vectors $\theta_2$ and $\theta_3$, respectively. \\
\noindent In this work, $y_i$ is tailored for a categorical classification problem.

\noindent\textbf{Posterior Probability Distribution.}
Since a priori knowledge on the latent factor does not exist, we have to marginalize over it in order to learn the model parameters, $\theta_1$, $\theta_2$, and $\theta_3$ in Equation (\ref{eq:eq3}) and (\ref{eq:eq4}). The non-linear neural network functions make inference intractable. 
Hence, variational inference \cite{kingma2013auto} along with the posterior network is employed.
These neural networks output the parameters of a fixed form posterior approximation over the latent variable $\textbf{z}$, given the observed variables. 
Similar to \cite{louizos2017causal,rezende2014stochastic}, in this work, the proposed posterior network is conditioning on observations. Also, the true posterior over $\textbf{Z}$ depends on $\textbf{X}$, $\textbf{t}$ and $\textbf{y}$.
Hence, the posterior approximation defined below is employed to build the posterior network.
\begin{align*}
    q(\textbf{z}_i | \textbf{x}_i, y_i, t_i) = \prod_{z\in \textbf{z}_i} \mathcal{N}(z | \bm{\mu}_i, \bm{\sigma^2}_i)
\label{eq:eq5}
\end{align*}
\begin{equation}
    \bm{\mu}_i = t_i \bm{\mu}_{t=1, i} + (1 - t_i) \bm{\mu}_{t=0,i} \nonumber
    \label{equ_6}
\end{equation}
\begin{equation}
    \bm{\sigma^2}_i = t_i \bm{\sigma^2}_{t=1, i} + (1 - t_i) \bm{\sigma^2}_{t=0, i} \nonumber
    \label{equ_7}
\end{equation}
\begin{equation}
    \bm{\mu}_{t=0, i} = g_{\phi_1} \circ g_{\phi_0}(\textbf{x}_i, y_i) \nonumber
    \label{equ_8}
\end{equation}
\begin{equation}
    \bm{\sigma^2}_{t=0, i} = \sigma(g_{\phi_2} \circ g_{\phi_0}(\textbf{x}_i, y_i)) \nonumber
    \label{equ_9}
\end{equation}
\begin{equation}
    \bm{\mu}_{t=1, i} = g_{\phi_3} \circ g_{\phi_0}(\textbf{x}_i, y_i) \nonumber
    \label{equ_10}
\end{equation}
\begin{equation}
    \bm{\sigma^2}_{t=1, i} = \sigma(g_{\phi_4} \circ g_{\phi_0}(\textbf{x}_i, y_i)), \nonumber
    \label{equ_11}
\end{equation}
where $g_{\phi_k}(\cdot)$ denotes a neural network with learnable parameters $\phi_k$ for the indices $k=0, 1, 2, 3, 4$, and $g_{\phi_0}(\textbf{x}_i, y_i)$ is a shared representation.

\subsection{Training Objective with Helper Distributions}

\begin{figure}[t!]
\begin{center}
\includegraphics[width=1.0\linewidth]{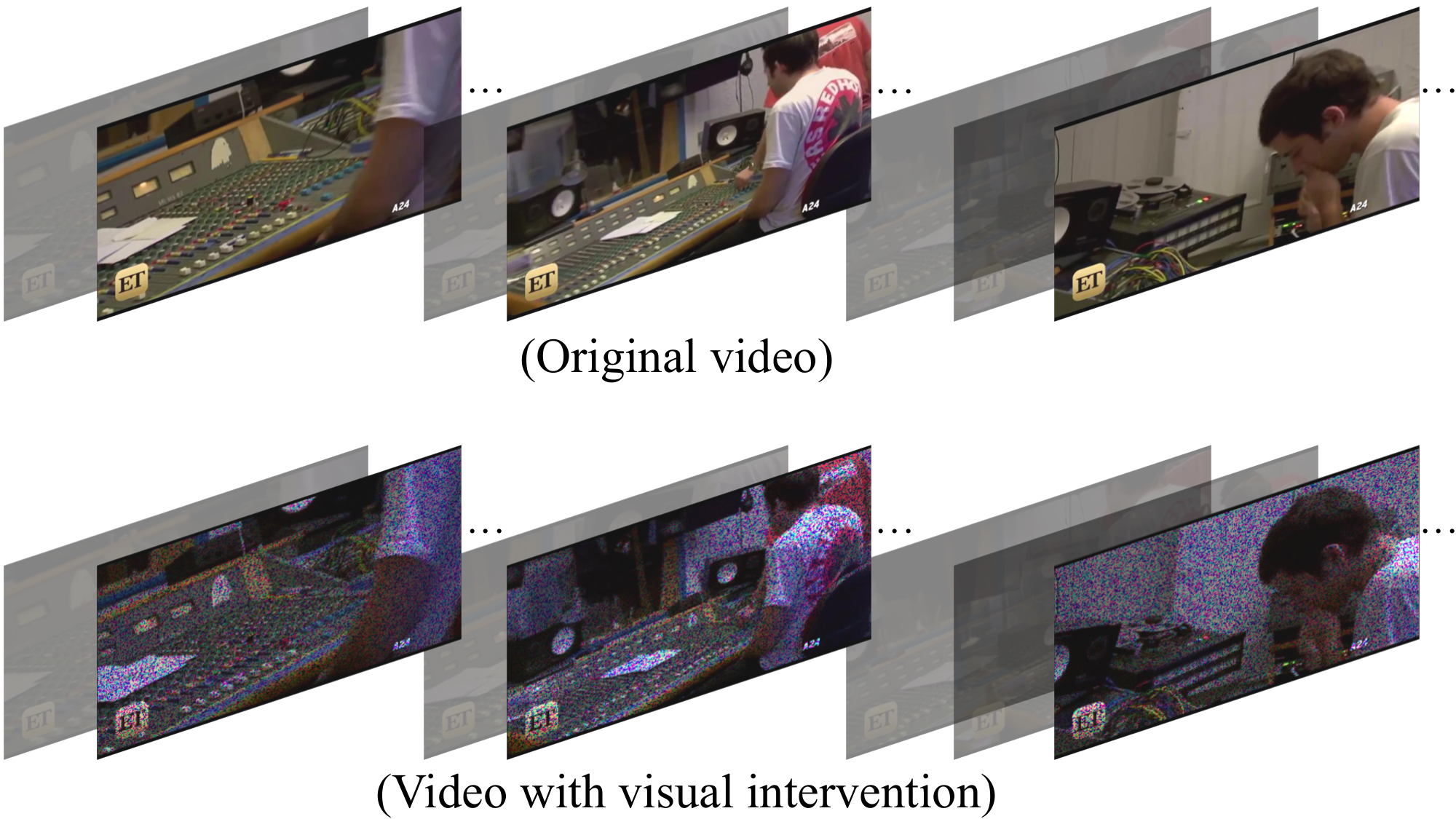}
\end{center}
\vspace{-0.50cm}
  \caption{A dataset example with the visual intervention of ``salt and pepper''. 
  }
\label{fig:figure5}
\end{figure}

\begin{figure}[t!]
\begin{center}
\includegraphics[width=1.0\linewidth]{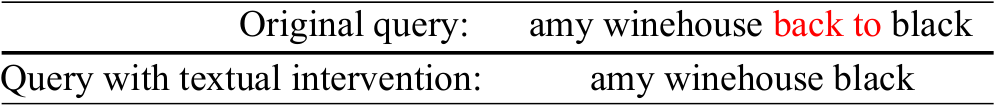}
\end{center}
\vspace{-0.50cm}
  \caption{A dataset example with the textual intervention of ``randomly missing some words in a sentence''. }
\label{fig:figure6}
\end{figure}

In practice, predicting the behaviors of data intervention and the model's outcome can be challenging due to various uncontrollable factors, such as video noise, motion blur, or lens blur. Two helper distributions are introduced to alleviate this issue.
We have to know the intervention $\textbf{t}$ along with its outcome $\textbf{y}$ before inferring the distribution over $\textbf{Z}$. Hence, the helper distribution defined below is introduced for the intervention $\textbf{t}$, referring to Figure and \ref{fig:figure5} Figure \ref{fig:figure6} for illustrations of the role of $\textbf{t}$.
\begin{align}
    q(t_i | \textbf{x}_i) = \textup{Bernoulli}(\sigma(g_{\phi_5}(\textbf{x}_i))).
\end{align}
The other helper distribution defined below is introduced for the outcome $y_{i}$.
\begin{align}
    q(y_i | \textbf{x}_i, t_i) = \sigma(t_{i} g_{\phi_6}(\textbf{x}_i) + (1 - t_{i}) g_{\phi_7}(\textbf{x}_i)),
\end{align}
where $g_{\phi_k}(\cdot)$ indicates a neural network with variational parameters $\phi_k$ for the indices $k=5, 6, 7$. 

\noindent The introduced helper distributions benefit the prediction of $t_i$ and $y_i$ for new samples. To estimate the variational parameters of the distributions $q(t_i | \textbf{x}_i)$ and $q(y_i | \textbf{x}_i, t_i)$, a helper objective function $\mathcal{L}_{\textup{helper}} = \sum^{N}_{i=1}[ \log q(t_i=t_i^{*} | \textbf{x}_i^{*}) + \log q(y_i=y_i^{*} | \textbf{x}_i^{*}, t_i^{*})]$ is introduced to the final training objective over $N$ data samples, where $\textbf{x}_i^*$, $t_i^*$ and $y_i^*$ are the observed values in the training set.
Finally, the overall training objective $\mathcal{L}_{\textup{conditional}}$ for the proposed method is defined below.
\begin{equation}
    \mathcal{L}_{\textup{conditional}} = \mathcal{L}_{\textup{helper}} ~ + \nonumber
\end{equation}
\vspace{-0.4cm}
\begin{equation}
    \sum^{N}_{i=1}\mathbb{E}_{q(\textbf{z}_{i}|\textbf{x}_{i}, t_i, y_i)}[\log p(\textbf{x}_i, t_i | \textbf{z}_i) ~ + \nonumber
\end{equation}
\begin{equation}
    \log p(y_i|t_i, \textbf{z}_i) + \log p(\textbf{z}_i) - \log q(\textbf{z}_i|\textbf{x}_i, t_i, y_i)].
    \label{equ_15}
\end{equation}



\subsection{Conditional Attention Module}
Since the textual input cannot always help the model performance because of the ineffective extraction of mutual information from the visual and textual inputs \cite{huang2020query,song2015tvsum}, a self-attention-based conditional attention module is introduced. The proposed extractor is built on top of transformer blocks \cite{vaswani2017attention}. Vanilla transformers exploit all of the tokens in each layer for attention computation. The design philosophy of the proposed conditional attention module is effectively using fewer but relatively informative tokens to compute attention maps. The computation of the vanilla attention matrix $\mathscr{A} \in \mathbb{R}^{n\times n}$ is based on the dot-product \cite{vaswani2017attention}:
\begin{align}
    \mathscr{A} &= \textup{softmax}\left ( \frac{\mathbf{Q}\mathbf{K}^\top}{\sqrt{d}} \right );
    \mathbf{Q} = \mathbf{T}\mathbf{W}_{q},
    \mathbf{K} = \mathbf{T}\mathbf{W}_{k},
    \label{eq:eq11}
\end{align}
where the query matrix $\mathbf{Q} \in \mathbb{R}^{n\times d}$ and key matrix $\mathbf{K} \in \mathbb{R}^{n\times d}$ are generated by the linear projection of the input token matrix $\mathbf{T} \in \mathbb{R}^{n\times d_{m}}$ based on the learnable weight matrices $\mathbf{W}_{q} \in \mathbb{R}^{d_{m}\times d}$ and $\mathbf{W}_{k} \in \mathbb{R}^{d_{m}\times d}$. $n$ indicates the total number of input tokens. $d$ represents the embedding dimension and $d_{m}$ denotes the dimension of an input token. 
The new value matrix $\mathbf{V}_{\textup{new}} \in \mathbb{R}^{n\times d}$ can be obtained via
\begin{align}
    \mathbf{V}_{\textup{new}} = \mathscr{A}\mathbf{V};
    \mathbf{V} = \mathbf{T}\mathbf{W}_{v},
    \label{eq:eq22}
\end{align}
where the value matrix $\mathbf{V} \in \mathbb{R}^{n\times d}$ and $\mathbf{W}_{v} \in \mathbb{R}^{d_{m}\times d}$.

In \cite{vaswani2017attention}, the vanilla attention matrix is based on the calculation of all the query-key pairs. However, to have better computational efficiency, in the proposed conditional attention module, only the top $\kappa$ most similar keys and values for each query are used to compute the conditional attention matrix. Similar to \cite{vaswani2017attention}, all the queries and keys are calculated by the dot-product. Then, the row-wise top $\kappa$ elements are used for the \textup{softmax} calculation. In the proposed conditional attention module, the value matrix $\mathbf{V}_{\kappa} \in \mathbb{R}^{n\times d}$ is defined as
\begin{align}
    \mathbf{V}_{\kappa} =\textup{softmax}\left (\tau _{\kappa} (\mathbf{\mathscr{A}}) \right ) \mathbf{V}_{\textup{new}}  \nonumber \\
    = \textup{softmax}\left (\tau _{\kappa} \left ( \frac{\mathbf{Q}\mathbf{K}^\top}{\sqrt{d}} \right )\right )\mathbf{V}_{\textup{new}},
    \label{eq:eq33}
\end{align}
where $\tau _{\kappa}(\cdot)$ denotes an operator for the row-wise top $\kappa$ elements selection. $\tau _{\kappa}(\cdot)$ is defined as:
\begin{align}
[\tau _{\kappa}(\mathbf{\mathscr{A}})]_{ij}=\begin{cases}
\mathscr{A}_{ij} &, \mathscr{A}_{ij}\in \text{top $\kappa$ elements at row~$i$} \\ 
-\infty  &, \text{ otherwise}.
\end{cases}
\label{eq:eq44}
\end{align}

Then, $\mathbf{V}_{\kappa}$ can be further used to generate $\mathbf{X}_{\textup{mul}}$, i.e.,  an output of the proposed conditional attention module. The procedure for calculating $\mathbf{X}_{\textup{mul}}$ is defined below.
\begin{equation}
    Z_{\textup{ta}} = \textup{\textup{TextAtten}}(\textup{FFN}(\textup{LayerNorm}(\mathbf{V}_{\kappa})),
    \label{eq:zta}
\end{equation}
where $\textup{LayerNorm}(\cdot)$ denotes a layer normalization, $\textup{FFN}(\cdot)$ indicates a feed forward network, and $\textup{TextAtten}(\cdot)$ denotes an element-wise multiplication-based textual attention mechanism.
\begin{equation}
    Z_{\textup{va}} = \textup{VisualAtten}(\textup{C3D}(\textbf{x})),
    \label{eq:zva}
\end{equation}
where $\textbf{x}$ denotes an input video, $\textup{C3D}(\cdot)$ indicates an operation of the spatial-temporal feature extraction, e.g., the 3D version of ResNet-34 \cite{he2016deep,hara2018can}, for the input video, and $\textup{VisualAtten}(\cdot)$ indicates a visual attention mechanism based on the element-wise multiplication.
\begin{equation}
    \mathbf{X}_{\textup{mul}} = \textup{FC}(Z_{\textup{ta}} \odot Z_{\textup{va}}),
    \label{eq:xmul}
\end{equation}
where $\odot$ denotes feature concatenation and $\textup{FC}(\cdot)$ indicates a fully connected layer. Note that the conditional attention module's output $\mathbf{X}_{\textup{mul}}$ is an input of the proposed posterior network based on the scheme of using multi-modal inputs.

Similar to the final step of video summary generation in \cite{huang2020query}, after the end-to-end training of the proposed conditional video summarization model is complete, the trained model can be used for video summary generation. Finally, based on the generated score labels, a set of video frames is selected from the original input video to form a final video summary. 

\section{Experiments}
In this section, the experimental setup and the used datasets are first described in detail. Then, the effectiveness of the proposed video summarization method is evaluated, analyzed, and compared with existing state-of-the-art methods. Finally, a conditional graph, as illustrated in Figure \ref{fig:figure39}, for video summarization is shown to demonstrate the improvement of modeling explainability.


\subsection{Experimental Setup and Datasets Preparation}
\label{section:section4-1}

\noindent\textbf{Experimental Setup.}
In this work, the following three scenarios are considered. First, in a fully-supervised scheme, a full set of data with human expert annotations, i.e., frame-level labels, are used to train the proposed model. Secondly, in a fully-supervised scenario with multi-modal input, the text-based query is considered as an additional input. Third, the authors of \cite{song2015tvsum} empirically find that a two-second segment length is appropriate for capturing video local context with good visual coherence. Hence, a video segment-level score is produced per two seconds based on given frame-level scores. The segment-level label can be considered as a type of weak label in a weakly-supervised learning scheme \cite{cai2018weakly,chen2019weakly,apostolidis2021video}.

\noindent\textbf{Video Summarization Datasets.}
To ensure a fair comparison with existing video summarization methods, we evaluate our proposed approach primarily on widely-used datasets, including TVSum \cite{song2015tvsum}, QueryVS \cite{huang2020query}, and SumMe \cite{gygli2014creating}.
The TVSum dataset, proposed by \cite{song2015tvsum}, consists of $50$ videos from various genres, including documentaries, how-to videos, news, egocentric videos, and vlogs. Each video has one corresponding title that can be used as a text-based query input. The dataset is annotated by $20$ crowd-workers per video.
The length of the video in TVSum is ranging from $2$ to $10$ minutes. The human expert frame-level importance score label in TVSum is ranging from $1$ to $5$. 
QueryVS, introduced by \cite{huang2020query}, is a larger dataset containing $190$ videos. In QueryVS, each video has frame-based annotations based on the one frame per second (fps) frame rate.
The video length in QueryVS is ranging from $2$ to $3$ minutes. The human expert frame-level importance score label in QueryVS is ranging from $0$ to $3$. Every video is retrieved based on a given text-based query. 
SumMe, proposed by \cite{gygli2014creating}, is a benchmark dataset consisting of $25$ videos. Each video is annotated with at least $15$ human summaries ($390$ in total), acquired in a controlled psychological experiment, providing an objective way to evaluate video summarization models and gain new insights into video summarization.
The video duration in SumMe is ranging from $1$ to $6$ minutes. In SumMe, the importance score annotated by human experts ranges from $0$ to $1$. Note that SumMe is not used for multi-modal video summarization. Hence, we do not have textual input when a model is evaluated on this dataset. 
Videos from these datasets are sampled at $1$ fps. The input image size is $224$ by $224$ with RGB channels. Every channel is normalized by standard deviation $=(0.2737, 0.2631, 0.2601)$ and mean $=(0.4280, 0.4106, 0.3589)$. PyTorch and NVIDIA TITAN Xp GPU are used for the implementation and to train models for $60$ epochs with $1e-6$ learning rate. The Adam optimizer is used \cite{kingma2014adam}, with hyper-parameters set as $\epsilon=1e-8$, $\beta_{1}=0.9$, and $\beta_{2}=0.999$. 
Large-scale query-based video summarization datasets with human expert annotations can be expensive to create, which is why commonly-used datasets like TVSum, QueryVS, and SumMe tend to be relatively small. To validate our proposed conditional modeling approach, we introduce three new video summarization datasets based on TVSum, QueryVS, and SumMe, respectively. \\

\noindent\textbf{Conditional Learning Dataset.}
\label{section:section3-5-1}
When researchers observe people's writing behaviors, they notice certain common occurrences, such as synonym replacement, accidentally missing some words in a sentence, and so on \cite{brand1985hot,shermis2014challenges}. Motivated by the above, we randomly pick up one of the behaviors, e.g., accidentally missing some words in a sentence, and write a textual intervention function to simulate it. Similarly, we know that when people make videos in their daily life, some visual disturbances may exist, e.g., salt and pepper noise, image masking, blurring, and so on. We also pick up some of them, e.g., blur and salt and pepper noise, and make a visual intervention function to do the simulation. Based on the controllable visual and textual  intervention simulation functions, we can make our conditional video summarization dataset with visual and textual interventions. The dataset is made based on the following steps. First, 50\% of the \textit{(video, query)} data pairs are randomly selected from the original training, validation, and testing sets. Secondly, for each selected video, $0$ or $1$ intervention labels are randomly assigned to $30$\% of the video frames and the corresponding queries. Figure \ref{fig:figure5} and Figure \ref{fig:figure6} are dataset examples with visual intervention and textual intervention, respectively. Note that in the real world, there are various possible disturbances. The aforementioned randomly selected visual and textual interventions are just a few of them. The other interventions can also be used in the proposed method.

\begin{table}[ht]
    \caption{Comparison with fully-supervised state-of-the-art methods. The proposed method performs the best on both datasets. Note that textual query input is not used in this experiment.
    }
\vspace{-0.2cm}
\centering
\begin{tabular}{c|ccc}
\toprule
\multicolumn{2}{c|}{\textbf{Fully-supervised Method}} & \textbf{TVSum}   &    \textbf{SumMe}    \\ 
\midrule
\multicolumn{2}{c|}{vsLSTM \cite{zhang2016video}}   & 54.2         & 37.6                                \\ 
\midrule
\multicolumn{2}{c|}{dppLSTM \cite{zhang2016video}}  & 54.7        & 38.6                       \\ 
\midrule
\multicolumn{2}{c|}{SASUM \cite{wei2018video}}    & 53.9       & 40.6                                           \\ 
\midrule
\multicolumn{2}{c|}{ActionRanking \cite{elfeki2019video}}   & 56.3       & 40.1                             \\ 
\midrule
\multicolumn{2}{c|}{H-RNN \cite{zhao2017hierarchical}}   & 57.7        & 41.1                                          \\ 
\midrule
\multicolumn{2}{c|}{DR-DSN$_{sup}$ \cite{zhou2018deep}}  & 58.1          & 42.1                                         \\ 
\midrule
\multicolumn{2}{c|}{PCDL$_{sup}$ \cite{zhao2019property}}    & 59.2       & 43.7                                    \\ 
\midrule
\multicolumn{2}{c|}{UnpairedVSN$_{psup}$ \cite{rochan2019video}}   & 56.1         & 48.0                                   \\ 
\midrule
\multicolumn{2}{c|}{SUM-FCN \cite{rochan2018video}}  & 56.8        & 47.5                                              \\ 
\midrule
\multicolumn{2}{c|}{SF-CVS \cite{huang2019novel_1}}    & 58.0         & 46.0                                              \\ 
\midrule
\multicolumn{2}{c|}{SASUM$_{fullysup}$ \cite{wei2018video}}   & 58.2         & 45.3                                \\ 
\midrule
\multicolumn{2}{c|}{A-AVS \cite{ji2019video}}    & 59.4       & 43.9                                             \\  
\midrule
\multicolumn{2}{c|}{CRSum \cite{yuan2019spatiotemporal}}   & 58.0       & 47.3                                       \\
\midrule
\multicolumn{2}{c|}{HSA-RNN \cite{zhao2018hsa}}  & 59.8           & 44.1                                                \\ 
\midrule
\multicolumn{2}{c|}{M-AVS \cite{ji2019video}}   & 61.0       & 44.4                                           \\ 
\midrule
\multicolumn{2}{c|}{ACGAN$_{sup}$ \cite{he2019unsupervised}}  & 59.4       & 47.2                                \\ 
\midrule
\multicolumn{2}{c|}{SUM-DeepLab \cite{rochan2018video}}    & 58.4         & 48.8                                  \\ 
\midrule
\multicolumn{2}{c|}{CSNet$_{sup}$ \cite{jung2019discriminative}}   & 58.5          & 48.6                               \\ 
\midrule
\multicolumn{2}{c|}{DASP \cite{ji2020deep}}  & 63.6        & 45.5                                    \\ 
\midrule
\multicolumn{2}{c|}{SMLD \cite{chu2019spatiotemporal}}    & 61.0          & 47.6       \\ 
\midrule
\multicolumn{2}{c|}{H-MAN \cite{liu2019learning}}   & 60.4          & 51.8        \\ 
\midrule
\multicolumn{2}{c|}{VASNet \cite{fajtl2018summarizing}}  & 61.4       & 49.7                                    \\ 
\midrule
\multicolumn{2}{c|}{iPTNet \cite{jiang2022joint}}    & 63.4    & 54.5                       \\ 
\midrule
\rowcolor{mygray} \multicolumn{2}{c|}{\textbf{Ours}}    & \textbf{67.5}        & \textbf{52.4}                           \\
\bottomrule
\end{tabular}
\label{table:table3}
\end{table}

\begin{table}[ht]
    \caption{Comparison with the multi-modal state-of-the-art. The proposed method outperforms the existing multi-modal approaches. `-' denotes unavailability from previous work. 
    }
\vspace{-0.2cm}
\centering
\begin{tabular}{c|ccc}
\toprule
\multicolumn{2}{c|}{\textbf{Multi-modal Method}} & \textbf{TVSum}   &    \textbf{QueryVS}    \\ 
\midrule
\multicolumn{2}{c|}{DSSE \cite{yuan2017video}}    & 57.0     & -    \\
\midrule
\multicolumn{2}{c|}{QueryVS \cite{huang2020query}}    & -      & 41.4  \\
\midrule
\multicolumn{2}{c|}{DQSN \cite{zhou2018video}}    & 58.6     & - \\
\midrule
\multicolumn{2}{c|}{GPT2MVS \cite{huang2021gpt2mvs}}    & -       & 54.8  \\
\midrule
\rowcolor{mygray} \multicolumn{2}{c|}{\textbf{Ours}}   & \textbf{68.2}     & \textbf{55.5} \\
\bottomrule
\end{tabular}
\label{table:table4}
\end{table}

\begin{table}[ht]
    \caption{Comparison with weakly-supervised state-of-the-art methods. The performance of the proposed approach is better than the existing weakly-supervised method.
    }
\vspace{-0.2cm}
\centering
\begin{tabular}{c|cc}
\toprule
\multicolumn{2}{c|}{\textbf{Weakly-supervised Method}} & \textbf{TVSum}      \\ 
\midrule
\multicolumn{2}{c|}{Random summary}  & 54.4         \\
\midrule
\multicolumn{2}{c|}{WS-HRL \cite{chen2019weakly}}  & 58.4    \\ 
\midrule
\rowcolor{mygray} \multicolumn{2}{c|}{\textbf{Ours}} & \textbf{66.9} \\ 
\bottomrule
\end{tabular}
\label{table:table5}
\end{table}


\begin{table}[t!]
    \caption{Ablation study of the proposed method for video summarization. ``w/o'' indicates a model without using that specific component. ``w/o CM'' indicates conditional modeling is used. ``w/o C3D'' denotes 2D CNN is used for video encoding. ``w/ BoW'' indicates bag-of-word is used for query embedding instead of the Conditional Attention Module (CAM). ``w/o CAM'' means the input textual query is not available. ``w/o Helper Dist.'' indicates helper distributions are not introduced.  $\mathbf{\Delta_{1}}$ and $\mathbf{\Delta_{2}}$ show the $F_{1}$-score difference, compared with the full model.
    }
\vspace{-0.2cm}
\centering
\begin{tabular}{c|ccccc}
\toprule
\multicolumn{2}{c|}{\textbf{Model}}  & \textbf{QueryVS}  & $\mathbf{\Delta_{1}}$ & \textbf{TVSum}  & $\mathbf{\Delta_{2}}$  \\ 
\midrule
\multicolumn{2}{c|}{w/o CM}  & 51.8 & -3.7  & 59.2  &  -9.0    \\
\midrule
\multicolumn{2}{c|}{w/o C3D}  & 53.8 & -1.7 & 63.7   &   -4.5  \\
\midrule
\multicolumn{2}{c|}{w/ BoW}  & 52.8 & -2.7 & 67.8   &  -0.4 \\
\midrule
\multicolumn{2}{c|}{w/o CAM}  & 52.3 & -3.2 & 67.5  &  -0.7 \\
\midrule
\multicolumn{2}{c|}{w/o Helper Dist.}  & 54.1 &  -1.4 & 66.7   & -1.5 \\
\midrule
\multicolumn{2}{c|}{w/o ConditionalAtten}  & 54.9 & -0.6 & 67.6   & -0.6  \\
\midrule
\multicolumn{2}{c|}{w/o VisualAtten}  & 55.1 & -0.4 & 67.9   &  -0.3 \\
\midrule
\multicolumn{2}{c|}{w/o TextAtten}  & 55.0 &  -0.5 & 67.7   & -0.5 \\
\midrule
\rowcolor{mygray} \multicolumn{2}{c|}{\textbf{Full model}} & \textbf{55.5} &  \textbf{0} & \textbf{68.2}  &  \textbf{0}  \\
\bottomrule
\end{tabular}
\label{table:table2}
\end{table}

\subsection{Evaluation and Analysis}
\label{section:section4-2}

\noindent\textbf{Evaluation protocol.}
Following existing works \cite{jiang2022joint,huang2021gpt2mvs,huang2020query,gygli2014creating,song2015tvsum}, we evaluate the proposed method under the same setting. TVSum, QueryVS, and SumMe datasets are randomly divided into five splits, respectively. For each of them, $80$\% of the dataset is used for training, and the remaining for evaluation. 
We run the models five times and report the averaged results \cite{jiang2022joint,huang2021gpt2mvs,huang2020query}. 
$F_{1}$-score \cite{hripcsak2005agreement,gygli2014creating,song2015tvsum,jiang2022joint} is adopted to measure the matching degree of the generated video summaries $\mathbb{S}_{i}$ and the ground-truth video summaries $\hat{\mathbb{S}}_{i}$ for video $ \mathbf{x}_{i}$. It is defined based on precision and recall. The precision $P$ and recall $R$ based on the temporal overlap between $\hat{\mathbb{S}}_{i}$ and $\mathbb{S}_{i}$ are calculated as follows:
\begin{align}
    P = \frac{|\mathbb{S}_{i} \cap \hat{\mathbb{S}}_{i}|}{|\mathbb{S}_{i}|}, R = \frac{|\mathbb{S}_{i} \cap \hat{\mathbb{S}}_{i}|}{|\hat{\mathbb{S}}_{i}|}, F_{1}=\frac{2PR}{P+R}.
\end{align}
For videos with multiple human-annotated video summaries, the calculation of metrics in \cite{jiang2022joint,huang2021gpt2mvs} is followed.

\begin{figure}[t!]
\begin{center}
\includegraphics[width=1.0\linewidth]{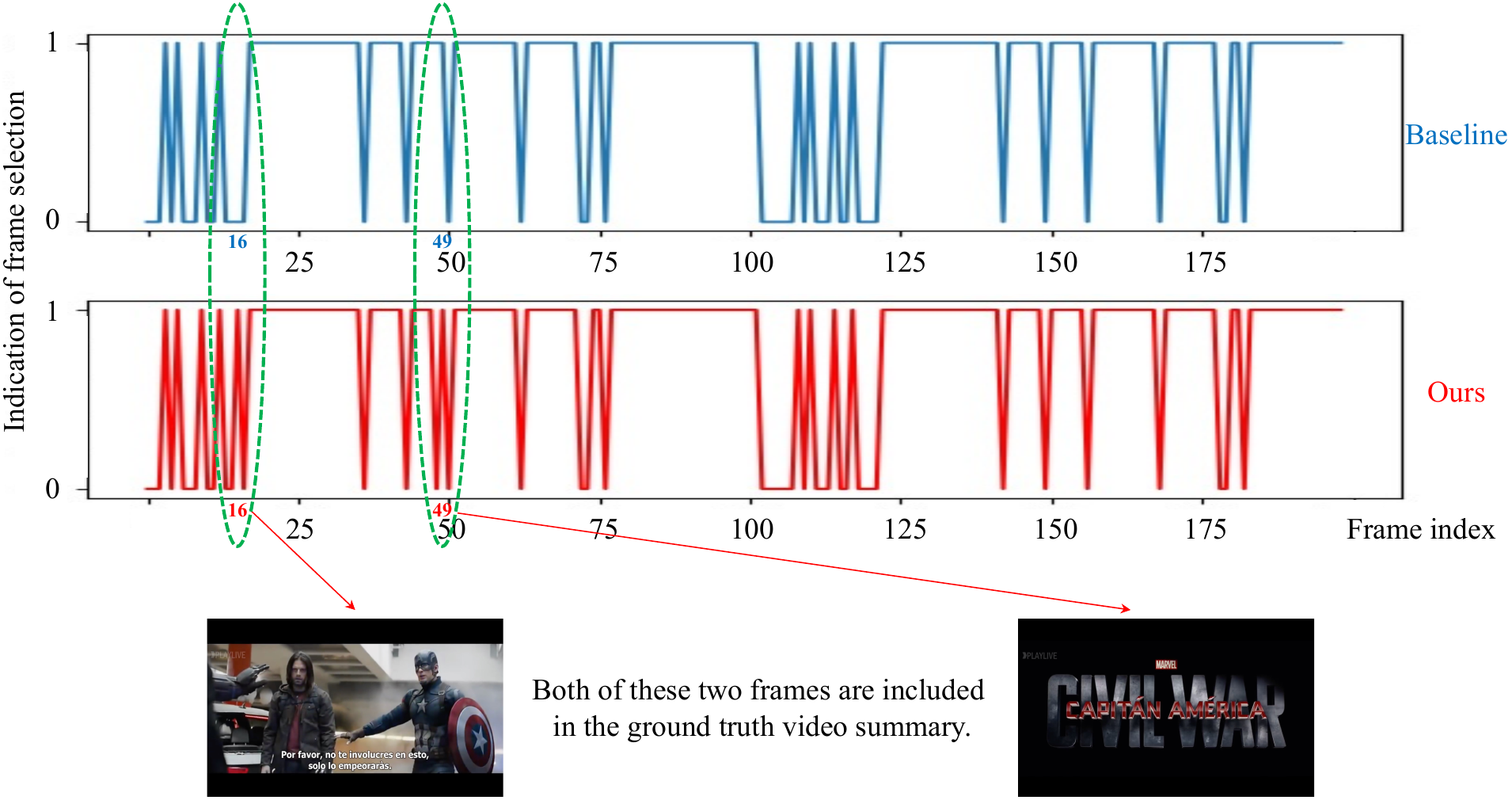}
\end{center}
\vspace{-0.50cm}
  \caption{This demonstration aims to show that the proposed method effectively considers visual and non-visual factors simultaneously. The comparison is conducted using the same baseline method and example video as shown in Figure \ref{fig:figure21}.
  }
\label{fig:figure399}
\end{figure}

\noindent\textbf{State-of-the-art comparisons.}
In Table \ref{table:table3}, Table \ref{table:table4}, Table \ref{table:table5}, and Figure \ref{fig:figure399}, the proposed method is compared with the existing state-of-the-art (SOTA) models based on the different supervision schemes. The results show that the proposed model beats the existing SOTA methods. The reason is that the introduced conditional modeling reinforces the conditional inference ability of a video summarization model. That is, it helps uncover the causal relations that steer the process and lead to the result.

\noindent\textbf{Ablation studies.}
To validate the effectiveness of the proposed method, the ablation study results are shown in Table \ref{table:table2}. The results show that the introduced components effectively boost the model performance. The reasons are as follows: (i) The proposed conditional attention module effectively captures the interaction in the various components of the proposed method. (ii) There exist explicit/implicit factors which will affect a video summarization model's conditional inference. Conditional attention effectively leverages the causal contribution of these factors. (iii) The contextualized query representation based on conditional attention is more effective than BoW \cite{scott1998text,soumya2014text} in terms of text-based query embedding. (iv) A text-based query is helpful to the model's performance. (v) A 3D CNN is more effective than a 2D CNN in terms of video encoding. (vi) Helper distributions improve the behavior prediction of the data intervention and the model’s outcome.

\noindent\textbf{Effectiveness analysis of the proposed conditional modeling.}
Since the main difference between the proposed approach and the existing methods is the proposed conditional modeling, the result in Table \ref{table:table3}, Table \ref{table:table4}, and Table \ref{table:table5} can be considered as an ablation study of the conditional learning based on different supervision schemes. The results show that the introduced conditional modeling is effective. One of the key components of the proposed method is the auxiliary task/distribution. The purpose of the introduced auxiliary task is to help the model learn to diagnose the input to make the correct inference of the main task, i.e., video summary inference, despite the irrelevant intervention that exists. In the training phase, the ground truth binary causation label is provided to teach the model whether the input has interfered or not. If a model can do well no matter whether the intervention exists or not, it means the model has the ability to analyze the input in order to perform well in the main task. In other words, it implies the model has conditional inference ability. In some sense, we also can say the proposed conditional modeling makes a model become more robust. 
However, while the aim of robustness analysis is to apply a very small intervention to an input to analyze a system \cite{huang2019novel,huang2017vqabq,huang2017robustness,huang2017robustnessMS,huang2023improving,huang2019assessing,huang2021deepopht,huang2022non,huang2021contextualized,huang2021deep,huang2021longer}, the input intervention in the proposed conditional modeling is intended to assist a model in learning the relationship between elements in a system. Therefore, the strength of an intervention in the proposed conditional modeling is not necessarily small, as the goal is to enhance the model's understanding of causal relationships.

\section{Discussion}
\noindent\textbf{Why video summarization can benefit from conditional modeling?}
1. Improved interpretability: Conditional modeling offers a framework to understand the relationships between different variables in a system, making it possible to build more interpretable video summarization models. By explicitly modeling the relationships between different elements of a video, such as objects, events, or scenes, we can gain better insights into the factors that contribute to the overall content and structure of the video summary. Since the proposed method is associated with a conditional graph of video summarization, the modeling explainability can benefit from the graph. In the introduced conditional video summarization model, we consider latent factors, which affect the generation of a good video summary, as the causal effect. Specifically, we exploit a conditional/causal graphical model to approach the video summarization problem. The modeling explainability in video summarization is illustrated in Figure \ref{fig:figure39}.

2. Better generalization: Conditional modeling can help in building more generalizable video summarization models that can work well across different video domains and contexts. By modeling the causal relationships between different variables, we can capture the underlying mechanisms that govern the behavior of the system, and use this knowledge to design more effective and robust video summarization models. 

3. Increased flexibility: Conditional modeling provides a flexible framework for building video summarization models that can be easily adapted to different scenarios and contexts. By explicitly modeling the relationships between different variables, we can modify and refine the model based on new observations or changes in the underlying data, without having to retrain the entire model from scratch. 

4. Better decision-making: Conditional modeling can help in building video summarization models that can make more informed and accurate decisions about what content to include in the summary. By modeling the causal relationships between different elements of the video, we can better understand how different factors (such as visual factors, semantic information, or temporal dynamics) contribute to the overall content and structure of the summary, and use this knowledge to make more informed decisions about what to include and what to leave out. 

\noindent\textbf{Main limitation of the proposed method.}
In general, annotated data is needed to train the proposed conditional model. Hence, the proposed approach cannot be considered an unsupervised video summarization method. The proposed method performs well, but the cost of annotated data cannot be ignored when deployed in the real world. In practice, the weakly-supervised model is a good way to balance the performance and deployment cost. See Table \ref{table:table5}.

\section{Conclusion}  
In conclusion, video summarization aims to automatically shorten videos while preserving the essential information required to convey the video's overall story. While existing methods mainly rely on visual factors like consecutiveness and diversity, they may overlook important non-visual factors such as interestingness, representativeness, and storyline consistency. This limitation can lead to suboptimal performance in generating high-quality video summaries.
To address these issues, a novel approach to video summarization was proposed in this work. Drawing insights from how humans create ground truth video summaries, the method employs a conditional modeling perspective and introduces meaningful random variables and joint distributions to capture key components of video summarization. Helper distributions are used to enhance model training, and a conditional attention module is designed to handle multi-modal input effectively. These innovative design choices aim to narrow the gap between human-generated and machine-generated video summaries.
The proposed video summarization method outperforms existing techniques and achieves state-of-the-art performance on commonly used video summarization datasets. Moreover, it enhances explainability as the conditional graph improves the understanding of decision-making in the model. This increased explainability is crucial for building trust in automated machine learning-based decision-making systems.
As video content continues to grow rapidly, the proposed video summarization method has the potential to significantly improve video exploration efficiency. By generating concise yet informative video summaries, it empowers users to efficiently process and comprehend vast amounts of video data, facilitating better decision-making and exploration in various applications.



%

\appendices


\ifCLASSOPTIONcompsoc
  \section*{Acknowledgments}
\else
  \section*{Acknowledgment}
\fi

This project has received funding from the European Union’s Horizon 2020 research and innovation programme under the Marie
Skłodowska-Curie grant agreement No 765140.

\ifCLASSOPTIONcaptionsoff
  \newpage
\fi



%
\bibliographystyle{unsrt}
\bibliography{deepeyenet}

\begin{thebibliography}{100}

\bibitem{plummer2017enhancing}
Bryan~A Plummer, Matthew Brown, and Svetlana Lazebnik.
\newblock Enhancing video summarization via vision-language embedding.
\newblock In {\em Proceedings of the IEEE conference on computer vision and pattern recognition}, pages 5781--5789, 2017.

\bibitem{chu2015video}
Wen-Sheng Chu, Yale Song, and Alejandro Jaimes.
\newblock Video co-summarization: Video summarization by visual co-occurrence.
\newblock In {\em Proceedings of the IEEE Conference on Computer Vision and Pattern Recognition}, pages 3584--3592, 2015.

\bibitem{panda2017collaborative}
Rameswar Panda and Amit~K Roy-Chowdhury.
\newblock Collaborative summarization of topic-related videos.
\newblock In {\em Proceedings of the IEEE Conference on Computer Vision and Pattern Recognition}, pages 7083--7092, 2017.

\bibitem{potapov2014category}
Danila Potapov, Matthijs Douze, Zaid Harchaoui, and Cordelia Schmid.
\newblock Category-specific video summarization.
\newblock In {\em ECCV}, pages 540--555. Springer, 2014.

\bibitem{rochan2019video}
Mrigank Rochan and Yang Wang.
\newblock Video summarization by learning from unpaired data.
\newblock In {\em Proceedings of the IEEE Conference on Computer Vision and Pattern Recognition}, pages 7902--7911, 2019.

\bibitem{li2018local}
Yandong Li, Liqiang Wang, Tianbao Yang, and Boqing Gong.
\newblock How local is the local diversity? reinforcing sequential determinantal point processes with dynamic ground sets for supervised video summarization.
\newblock In {\em ECCV}, pages 151--167, 2018.

\bibitem{zhou2018deep}
Kaiyang Zhou, Yu~Qiao, and Tao Xiang.
\newblock Deep reinforcement learning for unsupervised video summarization with diversity-representativeness reward.
\newblock In {\em Thirty-Second AAAI Conference on Artificial Intelligence}, 2018.

\bibitem{sharghi2018improving}
Aidean Sharghi, Ali Borji, Chengtao Li, Tianbao Yang, and Boqing Gong.
\newblock Improving sequential determinantal point processes for supervised video summarization.
\newblock In {\em ECCV}, pages 517--533, 2018.

\bibitem{zhang2018retrospective}
Ke~Zhang, Kristen Grauman, and Fei Sha.
\newblock Retrospective encoders for video summarization.
\newblock In {\em Proceedings of the European Conference on Computer Vision (ECCV)}, pages 383--399, 2018.

\bibitem{wu2022intentvizor}
Guande Wu, Jianzhe Lin, and Claudio~T Silva.
\newblock Intentvizor: Towards generic query guided interactive video summarization.
\newblock In {\em Proceedings of the IEEE/CVF Conference on Computer Vision and Pattern Recognition}, pages 10503--10512, 2022.

\bibitem{gygli2015video}
Michael Gygli, Helmut Grabner, and Luc Van~Gool.
\newblock Video summarization by learning submodular mixtures of objectives.
\newblock In {\em Proceedings of the IEEE conference on computer vision and pattern recognition}, pages 3090--3098, 2015.

\bibitem{song2015tvsum}
Yale Song, Jordi Vallmitjana, Amanda Stent, and Alejandro Jaimes.
\newblock Tvsum: Summarizing web videos using titles.
\newblock In {\em Proceedings of the IEEE conference on computer vision and pattern recognition}, pages 5179--5187, 2015.

\bibitem{vasudevan2017query}
Arun~Balajee Vasudevan, Michael Gygli, Anna Volokitin, and Luc Van~Gool.
\newblock Query-adaptive video summarization via quality-aware relevance estimation.
\newblock In {\em Proceedings of the 25th ACM international conference on Multimedia}, pages 582--590, 2017.

\bibitem{gong2014diverse}
Boqing Gong, Wei-Lun Chao, Kristen Grauman, and Fei Sha.
\newblock Diverse sequential subset selection for supervised video summarization.
\newblock In {\em Advances in neural information processing systems}, pages 2069--2077, 2014.

\bibitem{gygli2014creating}
Michael Gygli, Helmut Grabner, Hayko Riemenschneider, and Luc Van~Gool.
\newblock Creating summaries from user videos.
\newblock In {\em ECCV}, pages 505--520. Springer, 2014.

\bibitem{huang2020query}
Jia-Hong Huang and Marcel Worring.
\newblock Query-controllable video summarization.
\newblock In {\em ICMR}, pages 242--250, 2020.

\bibitem{jiang2022joint}
Hao Jiang and Yadong Mu.
\newblock Joint video summarization and moment localization by cross-task sample transfer.
\newblock In {\em Proceedings of the IEEE/CVF Conference on Computer Vision and Pattern Recognition}, pages 16388--16398, 2022.

\bibitem{huang2021gpt2mvs}
Jia-Hong Huang, Luka Murn, Marta Mrak, and Marcel Worring.
\newblock Gpt2mvs: Generative pre-trained transformer-2 for multi-modal video summarization.
\newblock In {\em ICMR}, pages 580--589, 2021.

\bibitem{louizos2017causal}
Christos Louizos, Uri Shalit, Joris~M Mooij, David Sontag, Richard Zemel, and Max Welling.
\newblock Causal effect inference with deep latent-variable models.
\newblock In {\em NIPS}, pages 6446--6456, 2017.

\bibitem{greenland1983correcting}
Sander Greenland and DAVID~G KLEINBAUM.
\newblock Correcting for misclassification in two-way tables and matched-pair studies.
\newblock {\em International Journal of Epidemiology}, 12(1):93--97, 1983.

\bibitem{selen1986adjusting}
Jan Sel{\'e}n.
\newblock Adjusting for errors in classification and measurement in the analysis of partly and purely categorical data.
\newblock {\em Journal of the American Statistical Association}, 81(393):75--81, 1986.

\bibitem{cai2012identifying}
Zhihong Cai and Manabu Kuroki.
\newblock On identifying total effects in the presence of latent variables and selection bias.
\newblock {\em arXiv preprint arXiv:1206.3239}, 2012.

\bibitem{greenland2011bias}
Sander Greenland and Timothy~L Lash.
\newblock Bias analysis.
\newblock {\em International Encyclopedia of Statistical Science}, 2:145--148, 2011.

\bibitem{shalit2017estimating}
Uri Shalit, Fredrik~D Johansson, and David Sontag.
\newblock Estimating individual treatment effect: generalization bounds and algorithms.
\newblock In {\em International Conference on Machine Learning}, pages 3076--3085. PMLR, 2017.

\bibitem{huang2022causal}
Jia-Hong Huang, Chao-Han~Huck Yang, Pin-Yu Chen, Andrew Brown, and Marcel Worring.
\newblock Causal video summarizer for video exploration.
\newblock In {\em 2022 IEEE International Conference on Multimedia and Expo (ICME)}, pages 1--6. IEEE, 2022.

\bibitem{zhang2016video}
Ke~Zhang, Wei-Lun Chao, Fei Sha, and Kristen Grauman.
\newblock Video summarization with long short-term memory.
\newblock In {\em ECCV}, pages 766--782. Springer, 2016.

\bibitem{zhao2017hierarchical}
Bin Zhao, Xuelong Li, and Xiaoqiang Lu.
\newblock Hierarchical recurrent neural network for video summarization.
\newblock In {\em Proceedings of the 25th ACM international conference on Multimedia}, pages 863--871, 2017.

\bibitem{zhao2018hsa}
Bin Zhao, Xuelong Li, and Xiaoqiang Lu.
\newblock Hsa-rnn: Hierarchical structure-adaptive rnn for video summarization.
\newblock In {\em Proceedings of the IEEE Conference on Computer Vision and Pattern Recognition}, pages 7405--7414, 2018.

\bibitem{zhang2019dtr}
Yujia Zhang, Michael Kampffmeyer, Xiaoguang Zhao, and Min Tan.
\newblock Dtr-gan: Dilated temporal relational adversarial network for video summarization.
\newblock In {\em Proceedings of the ACM Turing Celebration Conference-China}, pages 1--6, 2019.

\bibitem{ji2019video}
Zhong Ji, Kailin Xiong, Yanwei Pang, and Xuelong Li.
\newblock Video summarization with attention-based encoder-decoder networks.
\newblock {\em IEEE Transactions on Circuits and Systems for Video Technology}, 2019.

\bibitem{ji2020deep}
Zhong Ji, Fang Jiao, Yanwei Pang, and Ling Shao.
\newblock Deep attentive and semantic preserving video summarization.
\newblock {\em Neurocomputing}, 405:200--207, 2020.

\bibitem{hochreiter1997long}
Sepp Hochreiter and J{\"u}rgen Schmidhuber.
\newblock Long short-term memory.
\newblock {\em Neural computation}, 9(8):1735--1780, 1997.

\bibitem{kulesza2012determinantal}
Alex Kulesza and Ben Taskar.
\newblock Determinantal point processes for machine learning.
\newblock {\em arXiv preprint arXiv:1207.6083}, 2012.

\bibitem{li2017extracting}
Yujie Li, Atsunori Kanemura, Hideki Asoh, Taiki Miyanishi, and Motoaki Kawanabe.
\newblock Extracting key frames from first-person videos in the common space of multiple sensors.
\newblock In {\em 2017 IEEE International Conference on Image Processing (ICIP)}, pages 3993--3997. IEEE, 2017.

\bibitem{sanabria2019deep}
Melissa Sanabria, Fr{\'e}d{\'e}ric Precioso, and Thomas Menguy.
\newblock A deep architecture for multimodal summarization of soccer games.
\newblock In {\em Proceedings Proceedings of the 2nd International Workshop on Multimedia Content Analysis in Sports}, pages 16--24, 2019.

\bibitem{song2016category}
Xinhui Song, Ke~Chen, Jie Lei, Li~Sun, Zhiyuan Wang, Lei Xie, and Mingli Song.
\newblock Category driven deep recurrent neural network for video summarization.
\newblock In {\em 2016 IEEE International Conference on Multimedia \& Expo Workshops (ICMEW)}, pages 1--6. IEEE, 2016.

\bibitem{zhou2018video}
Kaiyang Zhou, Tao Xiang, and Andrea Cavallaro.
\newblock Video summarisation by classification with deep reinforcement learning.
\newblock {\em arXiv preprint arXiv:1807.03089}, 2018.

\bibitem{lei2018action}
Jie Lei, Qiao Luan, Xinhui Song, Xiao Liu, Dapeng Tao, and Mingli Song.
\newblock Action parsing-driven video summarization based on reinforcement learning.
\newblock {\em IEEE Transactions on Circuits and Systems for Video Technology}, 29(7):2126--2137, 2018.

\bibitem{otani2016video}
Mayu Otani, Yuta Nakashima, Esa Rahtu, Janne Heikkil{\"a}, and Naokazu Yokoya.
\newblock Video summarization using deep semantic features.
\newblock In {\em Asian Conference on Computer Vision}, pages 361--377. Springer, 2016.

\bibitem{yuan2017video}
Yitian Yuan, Tao Mei, Peng Cui, and Wenwu Zhu.
\newblock Video summarization by learning deep side semantic embedding.
\newblock {\em IEEE Transactions on Circuits and Systems for Video Technology}, 29(1):226--237, 2017.

\bibitem{wei2018video}
Huawei Wei, Bingbing Ni, Yichao Yan, Huanyu Yu, Xiaokang Yang, and Chen Yao.
\newblock Video summarization via semantic attended networks.
\newblock In {\em Proceedings of the AAAI Conference on Artificial Intelligence}, volume~32, 2018.

\bibitem{huang2023causalainer}
Jia-Hong Huang, Chao-Han~Huck Yang, Pin-Yu Chen, Min-Hung Chen, and Marcel Worring.
\newblock Causalainer: Causal explainer for automatic video summarization.
\newblock In {\em Proceedings of the IEEE/CVF Conference on Computer Vision and Pattern Recognition}, pages 2629--2635, 2023.

\bibitem{huang2023query}
Jia-Hong Huang, Luka Murn, Marta Mrak, and Marcel Worring.
\newblock Query-based video summarization with pseudo label supervision.
\newblock In {\em 2023 IEEE International Conference on Image Processing (ICIP)}, pages 1430--1434. IEEE, 2023.

\bibitem{hu2019silco}
Tao Hu, Pascal Mettes, Jia-Hong Huang, and Cees~GM Snoek.
\newblock Silco: Show a few images, localize the common object.
\newblock In {\em ICCV}, pages 5067--5076, 2019.

\bibitem{huck2018auto}
C-H Huck~Yang, Fangyu Liu, Jia-Hong Huang, Meng Tian, I-Hung Lin, Yi~Chieh Liu, Hiromasa Morikawa, Hao-Hsiang Yang, and Jesper Tegner.
\newblock Auto-classification of retinal diseases in the limit of sparse data using a two-streams machine learning model.
\newblock In {\em ACCV}, pages 323--338. Springer, 2018.

\bibitem{liu2018synthesizing}
Yi-Chieh Liu, Hao-Hsiang Yang, C-H Huck~Yang, Jia-Hong Huang, Meng Tian, Hiromasa Morikawa, Yi-Chang~James Tsai, and Jesper Tegner.
\newblock Synthesizing new retinal symptom images by multiple generative models.
\newblock In {\em ACCV}, pages 235--250. Springer, 2018.

\bibitem{yang2018novel}
C-H~Huck Yang, Jia-Hong Huang, Fangyu Liu, Fang-Yi Chiu, Mengya Gao, Weifeng Lyu, Jesper Tegner, et~al.
\newblock A novel hybrid machine learning model for auto-classification of retinal diseases.
\newblock {\em Workshop on Computational Biology, ICML}, 2018.

\bibitem{di2021dawn}
Riccardo Di~Sipio, Jia-Hong Huang, Samuel Yen-Chi Chen, Stefano Mangini, and Marcel Worring.
\newblock The dawn of quantum natural language processing.
\newblock {\em ICASSP}, 2022.

\bibitem{wu2023expert}
Ting-Wei Wu, Jia-Hong Huang, Joseph Lin, and Marcel Worring.
\newblock Expert-defined keywords improve interpretability of retinal image captioning.
\newblock In {\em Proceedings of the IEEE/CVF Winter Conference on Applications of Computer Vision}, pages 1859--1868, 2023.

\bibitem{panda2017weakly}
Rameswar Panda, Abir Das, Ziyan Wu, Jan Ernst, and Amit~K Roy-Chowdhury.
\newblock Weakly supervised summarization of web videos.
\newblock In {\em Proceedings of the IEEE International Conference on Computer Vision}, pages 3657--3666, 2017.

\bibitem{ho2018summarizing}
Hsuan-I Ho, Wei-Chen Chiu, and Yu-Chiang~Frank Wang.
\newblock Summarizing first-person videos from third persons' points of view.
\newblock In {\em Proceedings of the European Conference on Computer Vision (ECCV)}, pages 70--85, 2018.

\bibitem{cai2018weakly}
Sijia Cai, Wangmeng Zuo, Larry~S Davis, and Lei Zhang.
\newblock Weakly-supervised video summarization using variational encoder-decoder and web prior.
\newblock In {\em Proceedings of the European Conference on Computer Vision (ECCV)}, pages 184--200, 2018.

\bibitem{chen2019weakly}
Yiyan Chen, Li~Tao, Xueting Wang, and Toshihiko Yamasaki.
\newblock Weakly supervised video summarization by hierarchical reinforcement learning.
\newblock In {\em Proceedings of the ACM Multimedia Asia}, 2019.

\bibitem{jiang2019comprehensive}
Yudong Jiang, Kaixu Cui, Bo~Peng, and Changliang Xu.
\newblock Comprehensive video understanding: Video summarization with content-based video recommender design.
\newblock In {\em Proceedings of the IEEE/CVF International Conference on Computer Vision Workshops}, pages 0--0, 2019.

\bibitem{yan2020self}
Xiang Yan, Syed~Zulqarnain Gilani, Mingtao Feng, Liang Zhang, Hanlin Qin, and Ajmal Mian.
\newblock Self-supervised learning to detect key frames in videos.
\newblock {\em Sensors}, 20(23):6941, 2020.

\bibitem{zhao2014quasi}
Bin Zhao and Eric~P Xing.
\newblock Quasi real-time summarization for consumer videos.
\newblock In {\em Proceedings of the IEEE conference on computer vision and pattern recognition}, pages 2513--2520, 2014.

\bibitem{mahasseni2017unsupervised}
Behrooz Mahasseni, Michael Lam, and Sinisa Todorovic.
\newblock Unsupervised video summarization with adversarial lstm networks.
\newblock In {\em Proceedings of the IEEE conference on Computer Vision and Pattern Recognition}, pages 202--211, 2017.

\bibitem{herranz2012scalable}
Luis Herranz, Janko Calic, Jos{\'e}~M Mart{\'\i}nez, and Marta Mrak.
\newblock Scalable comic-like video summaries and layout disturbance.
\newblock {\em IEEE transactions on multimedia}, 14(4):1290--1297, 2012.

\bibitem{apostolidis2019stepwise}
Evlampios Apostolidis, Alexandros~I Metsai, Eleni Adamantidou, Vasileios Mezaris, and Ioannis Patras.
\newblock A stepwise, label-based approach for improving the adversarial training in unsupervised video summarization.
\newblock In {\em Proceedings of the 1st International Workshop on AI for Smart TV Content Production, Access and Delivery}, pages 17--25, 2019.

\bibitem{jung2019discriminative}
Yunjae Jung, Donghyeon Cho, Dahun Kim, Sanghyun Woo, and In~So Kweon.
\newblock Discriminative feature learning for unsupervised video summarization.
\newblock In {\em Proceedings of the AAAI Conference}, volume~33, pages 8537--8544, 2019.

\bibitem{yuan2019cycle}
Li~Yuan, Francis~EH Tay, Ping Li, Li~Zhou, and Jiashi Feng.
\newblock Cycle-sum: cycle-consistent adversarial lstm networks for unsupervised video summarization.
\newblock In {\em Proceedings of the AAAI Conference on Artificial Intelligence}, volume~33, pages 9143--9150, 2019.

\bibitem{apostolidis2020unsupervised}
Evlampios Apostolidis, Eleni Adamantidou, Alexandros~I Metsai, Vasileios Mezaris, and Ioannis Patras.
\newblock Unsupervised video summarization via attention-driven adversarial learning.
\newblock In {\em International Conference on Multimedia Modeling}, pages 492--504. Springer, 2020.

\bibitem{sheinfeld2016video}
Shay Sheinfeld, Yotam Gingold, and Ariel Shamir.
\newblock Video summarization using causality graphs.
\newblock In {\em HCOMP Workshop on Human Computation for Image and Video Analysis}, volume~2, 2016.

\bibitem{miao2018identifying}
Wang Miao, Zhi Geng, and Eric~J Tchetgen~Tchetgen.
\newblock Identifying causal effects with proxy variables of an unmeasured confounder.
\newblock {\em Biometrika}, 105(4):987--993, 2018.

\bibitem{pearl2012measurement}
Judea Pearl.
\newblock On measurement bias in causal inference.
\newblock {\em arXiv preprint arXiv:1203.3504}, 2012.

\bibitem{kuroki2014measurement}
Manabu Kuroki and Judea Pearl.
\newblock Measurement bias and effect restoration in causal inference.
\newblock {\em Biometrika}, 101(2):423--437, 2014.

\bibitem{wooldridge2009estimating}
Jeffrey~M Wooldridge.
\newblock On estimating firm-level production functions using proxy variables to control for unobservables.
\newblock {\em Economics letters}, 104(3):112--114, 2009.

\bibitem{edwards2015all}
Jessie~K Edwards, Stephen~R Cole, and Daniel Westreich.
\newblock All your data are always missing: incorporating bias due to measurement error into the potential outcomes framework.
\newblock {\em International journal of epidemiology}, 44(4):1452--1459, 2015.

\bibitem{abbasnejad2020counterfactual}
Ehsan Abbasnejad, Damien Teney, Amin Parvaneh, Javen Shi, and Anton van~den Hengel.
\newblock Counterfactual vision and language learning.
\newblock In {\em Proceedings of the IEEE/CVF Conference on Computer Vision and Pattern Recognition}, pages 10044--10054, 2020.

\bibitem{yang2021causal}
Chao-Han~Huck Yang, I~Hung, Te~Danny, Yi~Ouyang, and Pin-Yu Chen.
\newblock Causal inference q-network: Toward resilient reinforcement learning.
\newblock {\em arXiv preprint arXiv:2102.09677}, 2021.

\bibitem{agarwal2020towards}
Vedika Agarwal, Rakshith Shetty, and Mario Fritz.
\newblock Towards causal vqa: Revealing and reducing spurious correlations by invariant and covariant semantic editing.
\newblock In {\em Proceedings of the IEEE/CVF Conference on Computer Vision and Pattern Recognition}, pages 9690--9698, 2020.

\bibitem{pearl2018theoretical}
Judea Pearl.
\newblock Theoretical impediments to machine learning with seven sparks from the causal revolution.
\newblock {\em arXiv preprint arXiv:1801.04016}, 2018.

\bibitem{zhang2018advances}
Cheng Zhang, Judith B{\"u}tepage, Hedvig Kjellstr{\"o}m, and Stephan Mandt.
\newblock Advances in variational inference.
\newblock {\em IEEE transactions on pattern analysis and machine intelligence}, 41(8):2008--2026, 2018.

\bibitem{kingma2013auto}
Diederik~P Kingma and Max Welling.
\newblock Auto-encoding variational bayes.
\newblock {\em arXiv preprint arXiv:1312.6114}, 2013.

\bibitem{rezende2014stochastic}
Danilo~Jimenez Rezende, Shakir Mohamed, and Daan Wierstra.
\newblock Stochastic backpropagation and approximate inference in deep generative models.
\newblock {\em arXiv preprint arXiv:1401.4082}, 2014.

\bibitem{vaswani2017attention}
Ashish Vaswani, Noam Shazeer, Niki Parmar, Jakob Uszkoreit, Llion Jones, Aidan~N Gomez, Lukasz Kaiser, and Illia Polosukhin.
\newblock Attention is all you need.
\newblock {\em arXiv preprint arXiv:1706.03762}, 2017.

\bibitem{he2016deep}
Kaiming He, Xiangyu Zhang, Shaoqing Ren, and Jian Sun.
\newblock Deep residual learning for image recognition.
\newblock In {\em Proceedings of the IEEE conference on computer vision and pattern recognition}, pages 770--778, 2016.

\bibitem{hara2018can}
Kensho Hara, Hirokatsu Kataoka, and Yutaka Satoh.
\newblock Can spatiotemporal 3d cnns retrace the history of 2d cnns and imagenet?
\newblock In {\em Proceedings of the IEEE conference on Computer Vision and Pattern Recognition}, pages 6546--6555, 2018.

\bibitem{apostolidis2021video}
Evlampios Apostolidis, Eleni Adamantidou, Alexandros~I Metsai, Vasileios Mezaris, and Ioannis Patras.
\newblock Video summarization using deep neural networks: A survey.
\newblock {\em arXiv preprint arXiv:2101.06072}, 2021.

\bibitem{kingma2014adam}
Diederik~P Kingma and Jimmy Ba.
\newblock Adam: A method for stochastic optimization.
\newblock {\em arXiv preprint arXiv:1412.6980}, 2014.

\bibitem{brand1985hot}
Alice~G Brand.
\newblock Hot cognition: Emotions and writing behavior.
\newblock {\em Journal of advanced composition}, pages 5--15, 1985.

\bibitem{shermis2014challenges}
Mark~D Shermis.
\newblock The challenges of emulating human behavior in writing assessment.
\newblock {\em Assessing Writing}, 22:91--99, 2014.

\bibitem{elfeki2019video}
Mohamed Elfeki and Ali Borji.
\newblock Video summarization via actionness ranking.
\newblock In {\em 2019 IEEE Winter Conference on Applications of Computer Vision (WACV)}, pages 754--763. IEEE, 2019.

\bibitem{zhao2019property}
Bin Zhao, Xuelong Li, and Xiaoqiang Lu.
\newblock Property-constrained dual learning for video summarization.
\newblock {\em IEEE transactions on neural networks and learning systems}, 31(10):3989--4000, 2019.

\bibitem{rochan2018video}
Mrigank Rochan, Linwei Ye, and Yang Wang.
\newblock Video summarization using fully convolutional sequence networks.
\newblock In {\em ECCV}, pages 347--363, 2018.

\bibitem{huang2019novel_1}
Cheng Huang and Hongmei Wang.
\newblock A novel key-frames selection framework for comprehensive video summarization.
\newblock {\em IEEE Transactions on Circuits and Systems for Video Technology}, 30(2):577--589, 2019.

\bibitem{yuan2019spatiotemporal}
Yuan Yuan, Haopeng Li, and Qi~Wang.
\newblock Spatiotemporal modeling for video summarization using convolutional recurrent neural network.
\newblock {\em IEEE Access}, 7:64676--64685, 2019.

\bibitem{he2019unsupervised}
Xufeng He, Yang Hua, Tao Song, Zongpu Zhang, Zhengui Xue, Ruhui Ma, Neil Robertson, and Haibing Guan.
\newblock Unsupervised video summarization with attentive conditional generative adversarial networks.
\newblock In {\em Proceedings of the 27th ACM International Conference on Multimedia}, pages 2296--2304, 2019.

\bibitem{chu2019spatiotemporal}
Wei-Ta Chu and Yu-Hsin Liu.
\newblock Spatiotemporal modeling and label distribution learning for video summarization.
\newblock In {\em 2019 IEEE 21st International Workshop on Multimedia Signal Processing (MMSP)}, pages 1--6. IEEE, 2019.

\bibitem{liu2019learning}
Yen-Ting Liu, Yu-Jhe Li, Fu-En Yang, Shang-Fu Chen, and Yu-Chiang~Frank Wang.
\newblock Learning hierarchical self-attention for video summarization.
\newblock In {\em 2019 IEEE International Conference on Image Processing (ICIP)}, pages 3377--3381. IEEE, 2019.

\bibitem{fajtl2018summarizing}
Jiri Fajtl, Hajar~Sadeghi Sokeh, Vasileios Argyriou, Dorothy Monekosso, and Paolo Remagnino.
\newblock Summarizing videos with attention.
\newblock In {\em Asian Conference on Computer Vision}, pages 39--54. Springer, 2018.

\bibitem{hripcsak2005agreement}
George Hripcsak and Adam~S Rothschild.
\newblock Agreement, the f-measure, and reliability in information retrieval.
\newblock {\em Journal of the American medical informatics association}, 12(3):296--298, 2005.

\bibitem{scott1998text}
Sam Scott and Stan Matwin.
\newblock Text classification using wordnet hypernyms.
\newblock In {\em Usage of WordNet in Natural Language Processing Systems}, 1998.

\bibitem{soumya2014text}
K~Soumya~George and Shibily Joseph.
\newblock Text classification by augmenting bag of words (bow) representation with co-occurrence feature.
\newblock {\em IOSR J. Comput. Eng}, 16(1):34--38, 2014.

\bibitem{huang2019novel}
Jia-Hong Huang, Cuong~Duc Dao, Modar Alfadly, and Bernard Ghanem.
\newblock A novel framework for robustness analysis of visual qa models.
\newblock In {\em Proceedings of the Thirty-Third AAAI Conference on Artificial Intelligence}, pages 8449--8456, 2019.

\bibitem{huang2017vqabq}
Jia-Hong Huang, Modar Alfadly, and Bernard Ghanem.
\newblock Vqabq: Visual question answering by basic questions.
\newblock {\em VQA Challenge Workshop, CVPR}, 2017.

\bibitem{huang2017robustness}
Jia-Hong Huang, Modar Alfadly, and Bernard Ghanem.
\newblock Robustness analysis of visual qa models by basic questions.
\newblock {\em VQA Challenge and Visual Dialog Workshop, CVPR}, 2018.

\bibitem{huang2017robustnessMS}
Jia-Hong Huang.
\newblock Robustness analysis of visual question answering models by basic questions.
\newblock {\em King Abdullah University of Science and Technology, Master Thesis}, 2017.

\bibitem{huang2023improving}
Jia-Hong Huang, Modar Alfadly, Bernard Ghanem, and Marcel Worring.
\newblock Improving visual question answering models through robustness analysis and in-context learning with a chain of basic questions.
\newblock {\em arXiv preprint arXiv:2304.03147}, 2023.

\bibitem{huang2019assessing}
Jia-Hong Huang, Modar Alfadly, Bernard Ghanem, and Marcel Worring.
\newblock Assessing the robustness of visual question answering.
\newblock {\em arXiv preprint arXiv:1912.01452}, 2019.

\bibitem{huang2021deepopht}
Jia-Hong Huang, C-H~Huck Yang, Fangyu Liu, Meng Tian, Yi-Chieh Liu, Ting-Wei Wu, I~Lin, Kang Wang, Hiromasa Morikawa, Hernghua Chang, et~al.
\newblock Deepopht: medical report generation for retinal images via deep models and visual explanation.
\newblock In {\em WACV}, pages 2442--2452, 2021.

\bibitem{huang2022non}
Jia-Hong Huang, Ting-Wei Wu, C-H~Huck Yang, Zenglin Shi, I~Lin, Jesper Tegner, Marcel Worring, et~al.
\newblock Non-local attention improves description generation for retinal images.
\newblock In {\em WACV}, pages 1606--1615, 2022.

\bibitem{huang2021contextualized}
Jia-Hong Huang, Ting-Wei Wu, and Marcel Worring.
\newblock Contextualized keyword representations for multi-modal retinal image captioning.
\newblock In {\em ICMR}, pages 645--652, 2021.

\bibitem{huang2021deep}
Jia-Hong Huang, Ting-Wei Wu, Chao-Han~Huck Yang, and Marcel Worring.
\newblock Deep context-encoding network for retinal image captioning.
\newblock In {\em ICIP}, pages 3762--3766. IEEE, 2021.

\bibitem{huang2021longer}
Jia-Hong Huang, Ting-Wei Wu, Chao-Han~Huck Yang, and Marcel Worring.
\newblock Longer version for" deep context-encoding network for retinal image captioning".
\newblock {\em arXiv preprint arXiv:2105.14538}, 2021.

\end{thebibliography}




%






\begin{IEEEbiography}[{\includegraphics[width=1in,height=1.25in,clip,keepaspectratio]{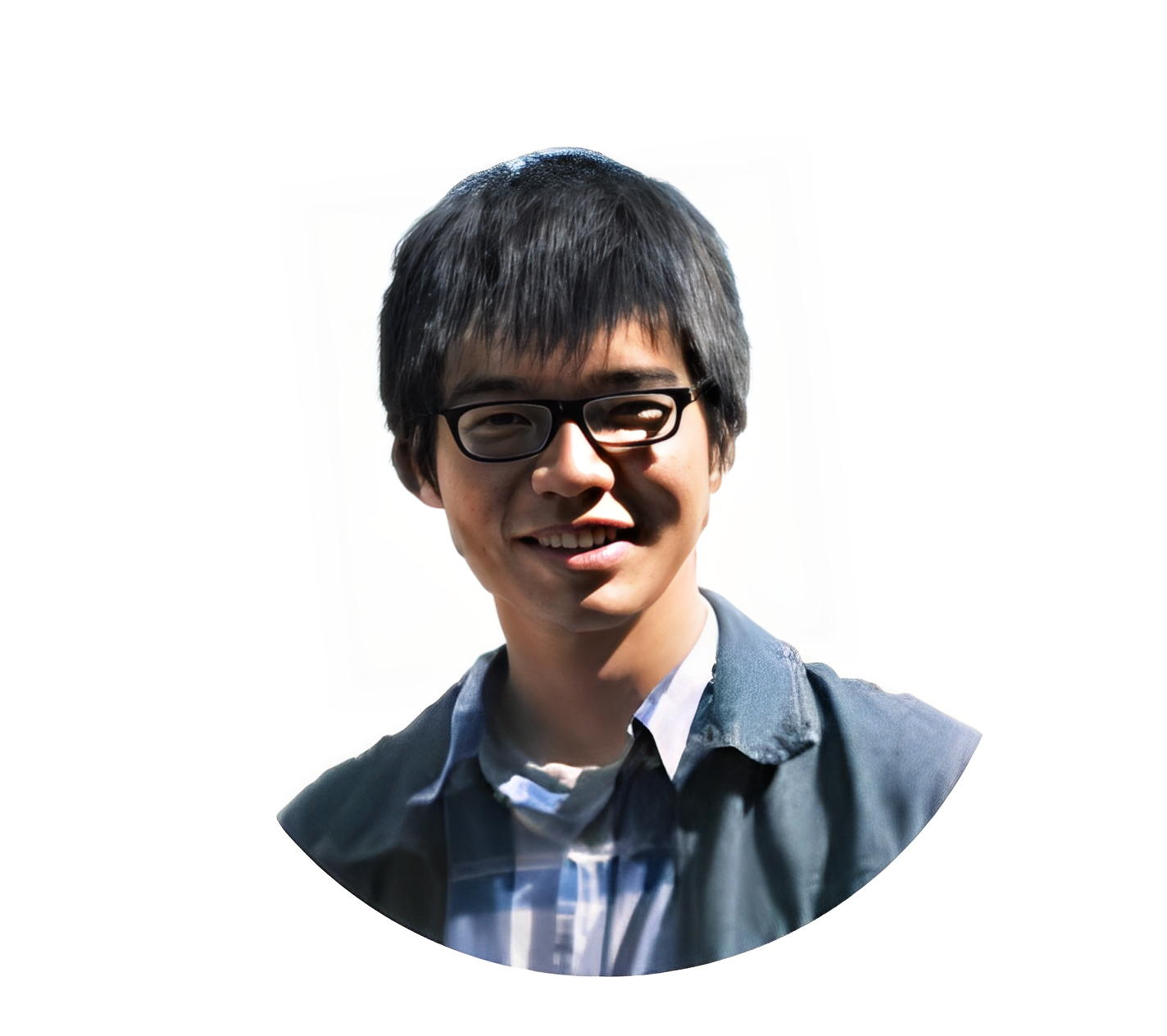}}]{Jia-Hong Huang}
Jia-Hong Huang, a final-year computer science Ph.D. student at the University of Amsterdam, holds Bachelor's degrees in Mathematical Sciences and Statistics from National Chengchi University (2014) and a Master's degree in Mechanical Engineering from National Taiwan University (2016). Additionally, he earned a Master's degree in Electrical Engineering from King Abdullah University of Science and Technology in 2018. His research focuses on multi-modal machine learning, specializing in algorithms for video, image, and text analysis. Jia-Hong has contributed significantly to automatic video summarization, medical report generation, and visual question answering. During his tenure at the University of Amsterdam, he has showcased prolific research output, publishing in renowned AI conferences such as AAAI, ICCV, CVPR, and more. Jia-Hong actively mentors students, organizes workshops, and has received awards, including the Government Scholarship to Study Abroad from Taiwan and the Marie Skłodowska–Curie Actions Fellowship from the European Union. Post-Ph.D., he aspires to pursue a career in academia or industry, contributing to cutting-edge research and mentoring future computer science researchers.

\end{IEEEbiography}

\begin{IEEEbiography}[{\includegraphics[width=1in,height=1.25in,clip,keepaspectratio]{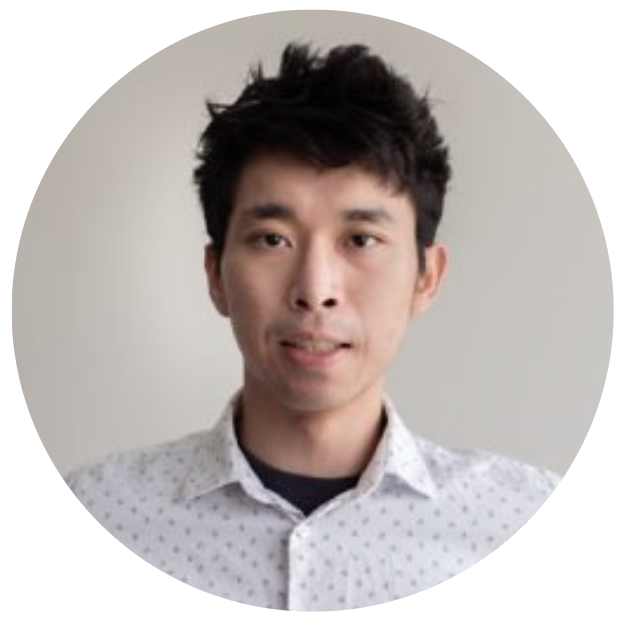}}]{Chao-Han Huck Yang}
received his Ph.D. and Master's degrees from the Georgia Institute of Technology in Atlanta, USA, and his Bachelor's degree from National Taiwan University. He worked as a research intern at Google, Amazon, and Hitachi, and his research interests include robust speech recognition, large-scale language modeling, parameter-efficient learning, and data privacy. Dr. Yang was an Area Chair for the Special Session on Sequence Modeling at ICASSP 2021 and presented tutorials at IJCAI, Interspeech, and ICASSP. He received the Outstanding Reviewer Award at NeurIPS 2021 and the Wallace H. Coulter Fellowship.

\end{IEEEbiography}

\begin{IEEEbiography}[{\includegraphics[width=1in,height=1.25in,clip,keepaspectratio]{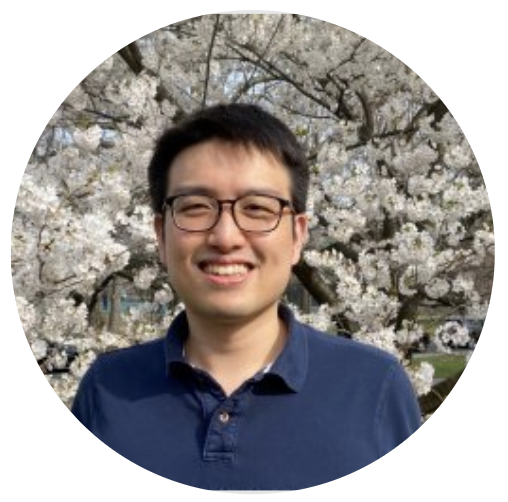}}]{Dr. Pin-Yu Chen}
is a principal research scientist at IBM Thomas J. Watson Research Center, Yorktown Heights, NY, USA. He is also the chief scientist of RPI-IBM AI Research Collaboration and PI of ongoing MIT-IBM Watson AI Lab projects. Dr. Chen received his Ph.D. in electrical engineering and computer science from the University of Michigan, Ann Arbor, USA, in 2016. Dr. Chen’s recent research focuses on adversarial machine learning and robustness of neural networks. His long-term research vision is to build trustworthy machine learning systems. He is a co-author of the book “Adversarial Robustness for Machine Learning”. At IBM Research, he received several research accomplishment awards, including IBM Master Inventor, IBM Corporate Technical Award, and IBM Pat Goldberg Memorial Best Paper. His research contributes to IBM open-source libraries including Adversarial Robustness Toolbox (ART 360) and AI Explainability 360 (AIX 360). He has published more than 50 papers related to trustworthy machine learning at major AI and machine learning conferences, given tutorials at NeurIPS’22, AAAI(’22,’23), IJCAI’21, CVPR(’20,’21,’23), ECCV’20, ICASSP(’20,’22,’23), KDD’19, and Big Data’18, and organized several workshops for adversarial machine learning. He received the IEEE GLOBECOM 2010 GOLD Best Paper Award and UAI 2022 Best Paper Runner-Up Award. 
\end{IEEEbiography}

\begin{IEEEbiography}[{\includegraphics[width=1in,height=1.25in,clip,keepaspectratio]{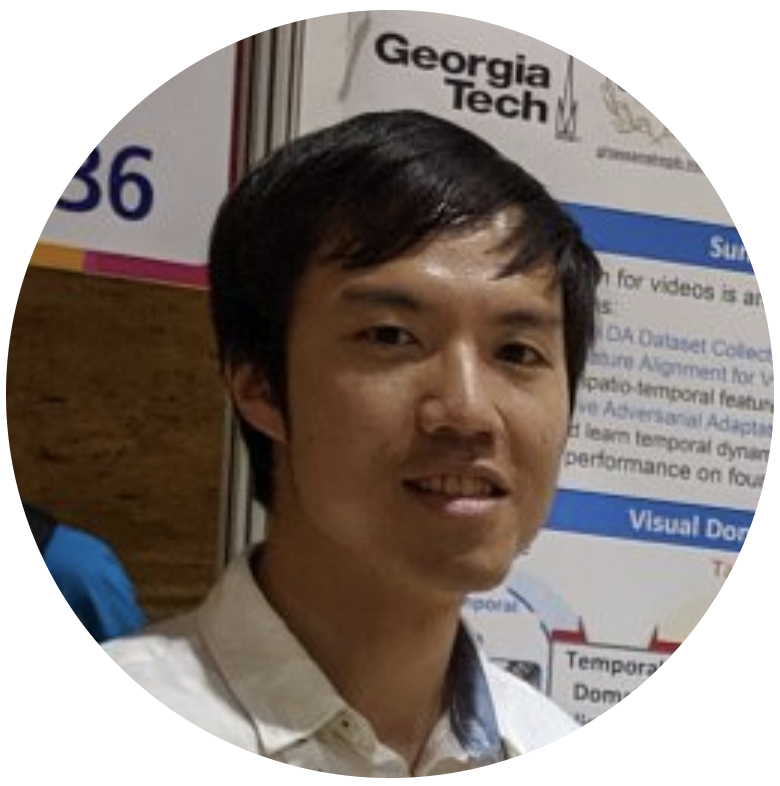}}]{Dr. Min-Hung Chen}
is a Research Scientist at NVIDIA Research, working on Vision+X Multi-Modal AI. He received his Ph.D. degree from Georgia Tech, advised by Prof. Ghassan AlRegib and in collaboration with Prof. Zsolt Kira. Before joining NVIDIA, Min-Hung was a Research Engineer II at Microsoft Azure AI, working on Cutting-edge AI Research for Cognitive Services. Before Microsoft, he was a Senior AI Engineer at MediaTek, working on Deep Learning Research for Edge-AI and Vision Transformer.
Min-Hung's research interest is mainly on Learning without Fully Supervision, including domain adaptation, continual learning, self-/semi-supervised learning, etc. In addition, he has also conducted research on transformer, attention, transfer learning, action segmentation, action recognition, video understanding, and temporal dynamics understanding. Please visit his website for more information.
\end{IEEEbiography}

\begin{IEEEbiography}[{\includegraphics[width=1in,height=1.25in,clip,keepaspectratio]{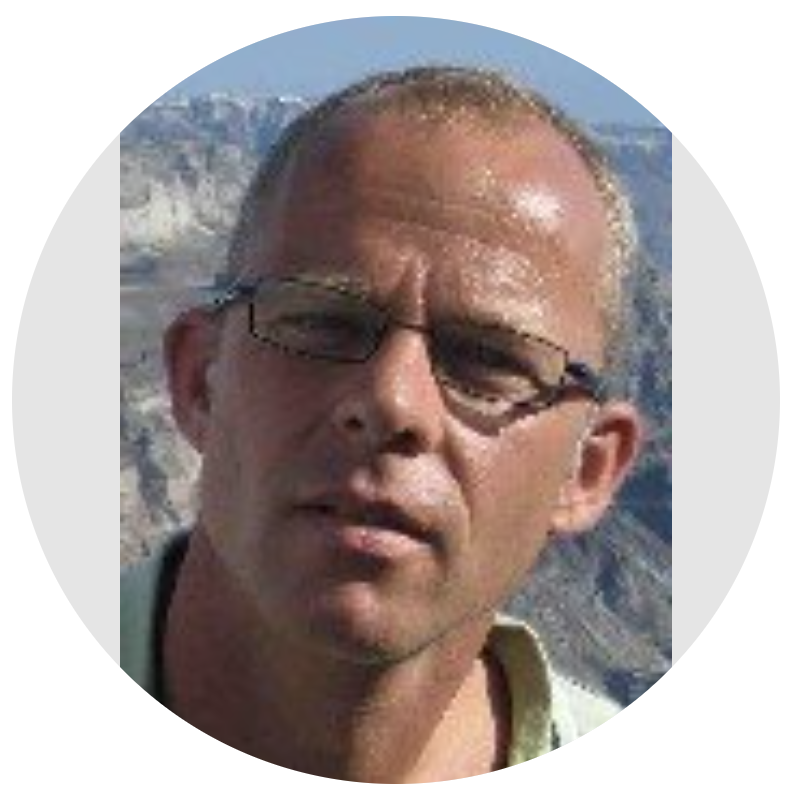}}]{Prof. Marcel Worring}
did his masters at the Free University Amsterdam in Computer Science with a specialization in medical computer science. During his Ph.D. work he focussed on image analysis both from a theoretical point of view, establishing the limits in accuracy one can achieve when measuring shape on a digital grid, to more applied shape analysis in biological and medical images. Part of this research was on deformable shape models and performed at Yale University. From there he moved to document and video analysis, spending four months in San Diego studying film theory and how it can help automatic video analysis, and subsequently more and more into the development of methods for accessing large image and video collections by their content. In such a setting, next to image analysis and understanding, user interaction is a crucial element and with it comes the need for advanced visualizations of the collection and the results of user queries. Currently he is taking this a step further into multimedia analytics being the integration of multimedia analysis, multimedia mining, information visualization, and multimedia interaction into a coherent framework yielding more than its constituent components.
\end{IEEEbiography}





\end{document}